\documentclass[pdflatex,sn-mathphys-num, iicol]{sn-jnl}

\usepackage{graphicx}%
\usepackage{multirow}%
\usepackage{amsmath,amssymb,amsfonts}%
\usepackage{amsthm}%
\usepackage{mathrsfs}%
\usepackage[title]{appendix}%
\usepackage{xcolor}%
\usepackage{textcomp}%
\usepackage{manyfoot}%
\usepackage{booktabs}%
\usepackage{algorithm}%
\usepackage{algorithmicx}%
\usepackage{algpseudocode}%
\usepackage{listings}%
\usepackage{rotating} 
\usepackage{stfloats}
\usepackage{float}

\usepackage[markup=underlined]{changes}

\theoremstyle{thmstyleone}%
\theoremstyle{thmstyletwo}%

\theoremstyle{thmstylethree}%

\begin{document}


\title[Article Title]{Generative AI for Misalignment-Resistant Virtual Staining to Accelerate Histopathology Workflows}


\author[1]{\fnm{Jiabo} \sur{Ma}}\email{jmabq@connect.ust.hk}
\equalcont{These authors contributed equally to this work.}
\author[1]{\fnm{Wenqiang} \sur{Li}}\email{wlidf@connect.ust.hk}
\equalcont{These authors contributed equally to this work.}
\author[2,3]{\fnm{Jinbang} \sur{Li}} \email{lzcy2008@126.com}
\author[1]{\fnm{Ziyi }\sur{Liu}}\email{cseziyiliu@ust.hk}
\author[1]{\fnm{Linshan} \sur{Wu}}\email{lwubf@connect.ust.hk}
\author[1]{\fnm{Fengtao} \sur{Zhou}} \email{fzhouaf@connect.ust.hk}
\author[2,3,4]{\fnm{Li} \sur{Liang}}\email{lli@smu.edu.cn}
\author[5]{\fnm{Ronald Cheong Kin} \sur{Chan}}\email{ronaldckchan@cuhk.edu.hk}
\author[6]{\fnm{Terence T. W.} \sur{Wong}}\email{ttwwong@ust.hk}
\author*[1,6,7,8,9]{\fnm{Hao} \sur{Chen}}\email{jhc@cse.ust.hk}

\affil[1]{\orgdiv{Department of Computer Science and Engineering}, \orgname{The Hong Kong University of Science and Technology}, \orgaddress{\state{Hong Kong SAR}, \country{China}}}
\affil[2]{\orgdiv{Department of Pathology, Nanfang Hospital and School of Basic Medical Sciences}, \orgname{Southern Medical University},
\orgaddress{\state{Guangzhou}, \country{China}}}
\affil[3]{\orgname{Guangdong Provincial Key Laboratory of Molecular Tumor Pathology},
\orgaddress{\state{Guangzhou}, \country{China}}}
\affil[4]{\orgname{Jinfeng Laboratory},
\orgaddress{\state{Chongqing}, \country{China}}}
\affil[5]{\orgdiv{Department of Anatomical and Cellular Pathology}, \orgname{The Chinese University of Hong Kong}, \orgaddress{\state{Hong Kong SAR}, \country{China}}}
\affil[6]{\orgdiv{Department of Chemical and Biological Engineering}, 
\orgname{The Hong Kong University of Science and Technology}, 
\orgaddress{\state{Hong Kong SAR}, \country{China}}}
\affil[7]{\orgdiv{Division of Life Science}, \orgname{The Hong Kong University of Science and Technology}, \orgaddress{\state{Hong Kong SAR}, \country{China}}}
\affil[8]{\orgdiv{State Key Laboratory of Nervous System Disorders}, \orgname{The Hong Kong University of Science and Technology}, \orgaddress{\state{Hong Kong SAR}, \country{China}}}
\affil[9]{\orgdiv{Shenzhen-Hong Kong Collaborative Innovation Research Institute}, \orgname{The Hong Kong University of Science and Technology}, \orgaddress{\state{Shenzhen}, \country{China}}}




\abstract{
Accurate histopathological diagnosis often requires multiple differently stained tissue sections, a process that is time-consuming, labor-intensive, and environmentally taxing due to the use of multiple chemical stains. Recently, virtual staining has emerged as a promising alternative that is faster, tissue-conserving, and environmentally friendly. However, existing virtual staining methods face significant challenges in clinical applications, primarily due to their reliance on well-aligned paired data. Obtaining such data is inherently difficult because chemical staining processes can distort tissue structures, and a single tissue section cannot undergo multiple staining procedures without damage or loss of information.
As a result, most available virtual staining datasets are either unpaired or roughly paired, making it difficult for existing methods to achieve accurate pixel-level supervision.
To address this challenge, we propose a robust virtual staining framework featuring cascaded registration mechanisms to resolve spatial mismatches between generated outputs and their corresponding ground truth.
Specifically, our method incorporates a registration module to correct alignment errors between generated images and corresponding roughly paired ground truth during the loss computation.
Additionally, we employ a separate registration network with adversarial training to fully decouple the generation and registration processes. 
This ensures that the generator focuses exclusively on image generation without the burden of handling alignment tasks. 
Importantly, our framework is designed to integrate seamlessly into existing virtual staining models without requiring modifications to their original network architectures, paving the way for further data and model scaling.
Experimental results demonstrate that our method significantly outperforms state-of-the-art models across five datasets, achieving an average improvement of 3.2\% on internal datasets and 10.1\% on external datasets. 
Moreover, in datasets with substantial misalignment, our approach achieves a remarkable 23.8\% improvement in peak signal-to-noise ratio compared to baseline models. 
Notably, a blinded evaluation by experienced pathologists revealed that they could not reliably distinguish between chemically stained and virtually stained images, with an accuracy of only 52\%, further emphasizing the potential of our proposed method.
The exceptional robustness of the proposed method across diverse datasets simplifies the data acquisition process for virtual staining and offers new insights for advancing its development.
}




\maketitle

\section{Introduction}\label{sec1}
\begin{figure*}[htbp]
    \centering
    \includegraphics[width=\textwidth]{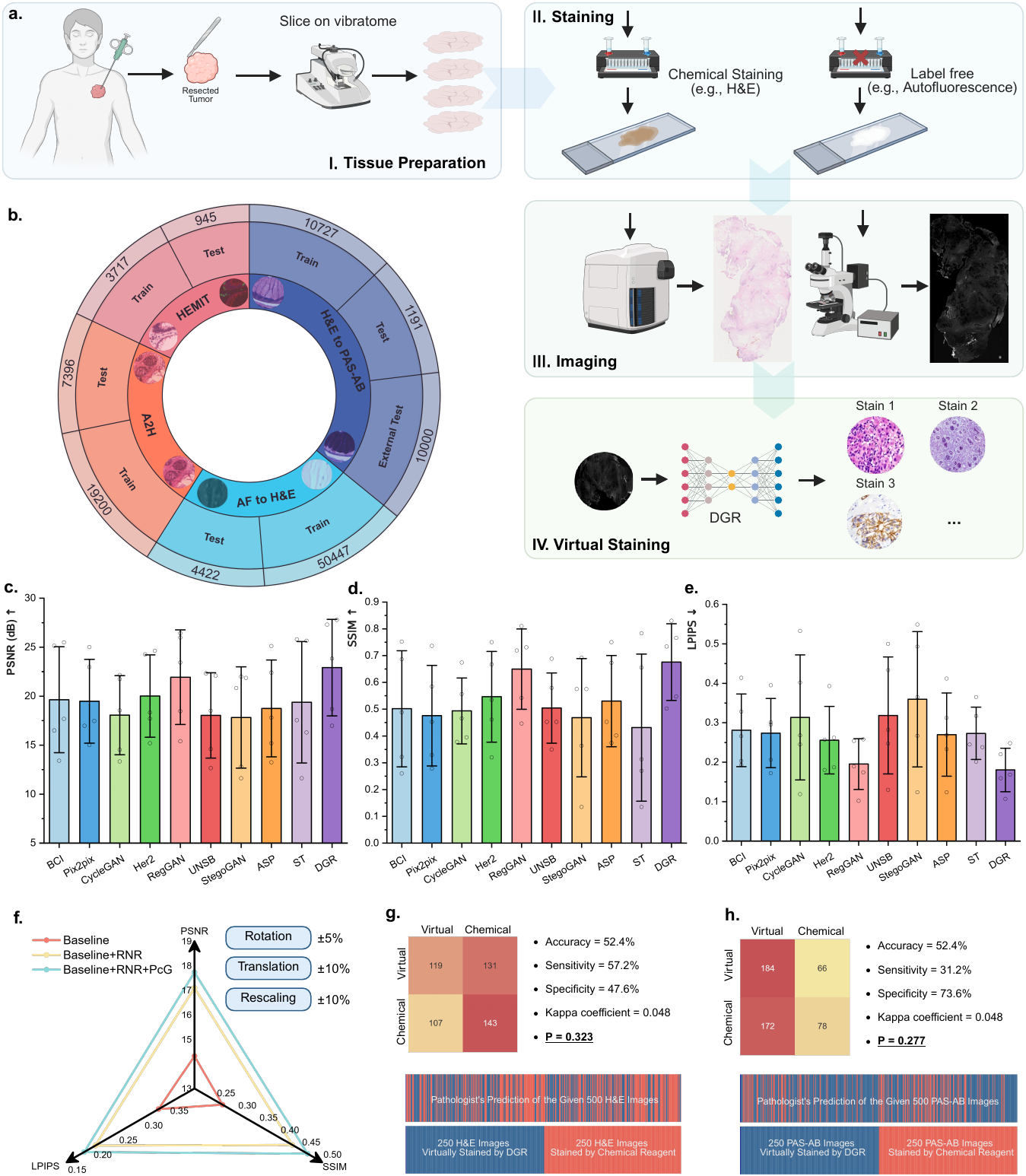}
    \caption{\textbf{Overview of virtual staining}. 
    \textbf{a.} The virtual staining workflow, illustrating the process from tissue sampling to virtually stained images. The model architecture of DGR is detailed in Extended Data Fig. \ref{fig:DTR-arch}.
    \textbf{b.} Datasets used to evaluate the performance of the DGR model. 
    \textbf{c-e.} Average performance comparisons of various virtual staining methods across five test sets. 
    Error bars represent standard deviations.
    \textbf{f.} Robustness comparison of different DGR modules under severe artificial misalignment  ( $\pm$5\% rotation, $\pm$10\% translation, and $\pm$10\% scaling). 
    \textbf{g-h.} Blinded evaluation by an experienced pathologist for distinguishing chemically and virtually stained H\&E and PAS-AB images, respectively.
The top section indicates the pathologist’s prediction, while the bottom section denotes the real staining method of the evaluated images. 
    }
    \label{fig1}
\end{figure*}
Histopathological analysis is a cornerstone of clinical diagnostics, enabling the identification and characterization of disease states based on the microscopic examination of tissue samples \cite{bancroft2008theory, musumeci2014past}.
Traditionally, this process relies on multiple chemically stained tissue sections, where each stain highlights specific tissue components or structures.
For example, hematoxylin and eosin (H\&E) staining, which differentiates cell nuclei from the extracellular tissue matrix, is the most widely used staining technique in histopathology \cite{titford2005long}. 
In contrast, the Periodic Acid-Schiff (PAS) stain, which specifically targets glycoproteins, is commonly utilized in renal pathology \cite{pollitt1996basement}.
However, the conventional staining pipeline is resource-intensive, requiring significant time, labor, and the use of chemical reagents that contribute to environmental burdens \cite{latonen2024virtual}.
These limitations have motivated the development of alternative approaches to streamline histopathological workflows, among which virtual staining has emerged as a promising solution \cite{, rivenson2019virtual, bai2023deep, pati2024accelerating}.
By digitally transforming label-free tissue images or one stain into target stain, virtual staining techniques offer the potential to save time, conserve tissue, and reduce dependency on chemical reagents, thereby addressing key challenges in modern diagnostic pathology (Fig. \ref{fig1}a).

Despite these advantages, virtual staining methods are not yet widely adopted in clinical settings due to significant technical challenges. 
Most existing methods rely on well-aligned paired datasets for training, which are difficult to obtain in practice \cite{ma2023efficient,lin2025virtual}.
Chemical staining processes inherently alter tissue structures, making it nearly impossible to achieve precise alignment between unstained and stained tissue sections \cite{li2024virtual, xu2025digital}. 
Furthermore, a single tissue section cannot undergo multiple staining procedures without causing damage or information loss, leaving most available datasets either unpaired or only roughly paired \cite{rivenson2019virtual}. 
This misalignment hinders the performance of current virtual staining methods, particularly for applications requiring pixel-level accuracy \cite{rivenson2019virtual, bai2023deep, pati2024accelerating, ma2023efficient, bai2022label, isola2017image}.
To address the problems caused by image misalignment, the Cycle-GAN framework \cite{zhu2017unpaired} has been widely adopted. 
This approach uses an additional generator and adopts cycle-consistency to train both generators, which usually perform better on unpaired data compared with Pix2Pix models \cite{isola2017image}. 
However, the cycle-consistency constraint can cause the model to focus more on image style than content, which may lead to content being lost in the generation process \cite{dubey2023structural, huang2023tc}.
Another strategy focuses on minimizing the influence of misalignment.
For instance, the pyramid Pix2Pix model \cite{Liu_2022_CVPR} incorporates a multi-scale loss function to enhance robustness against structural misalignment in the data.
Furthermore, RegGAN \cite{kong2021breaking} incorporates a registration model to align the generated image with the target image. 
By using the aligned generated image to compute the loss, RegGAN further reduces the impact of misalignment.
Ideally, the generated image should align well with the input image, while the registered generated image should closely match the target image. 
However, in RegGAN, only the latter condition is explicitly enforced because the generator and registration model are tightly coupled. 
This coupling lacks additional constraints to ensure accurate alignment between the input image and the generated image.

To address these limitations, we propose a robust virtual staining framework, termed DGR (Decoupled Generation and Registration), which decouples the image generation and registration processes. 
Specifically, we introduce two constraints: one to maintain alignment between the input image and the generated image, and another to ensure alignment between the registered generated image and the noisy ground truth. By decoupling the generator and registration networks, DGR enables effective virtual staining of roughly paired images.
We evaluated DGR against state-of-the-art virtual staining methods across five datasets. 
Experimental results demonstrate that DGR outperforms existing approaches, particularly in cases of severe misalignment.
Notably, DGR achieves a 3.4 dB (23.8\%) improvement over baseline methods. Furthermore, DGR significantly reduces the reliance on accurately aligned paired images, which is critical for real-world applications of virtual staining in pathology.
\section{Results}
\begin{figure*}[htbp]
    \centering
    \includegraphics[width=\textwidth]{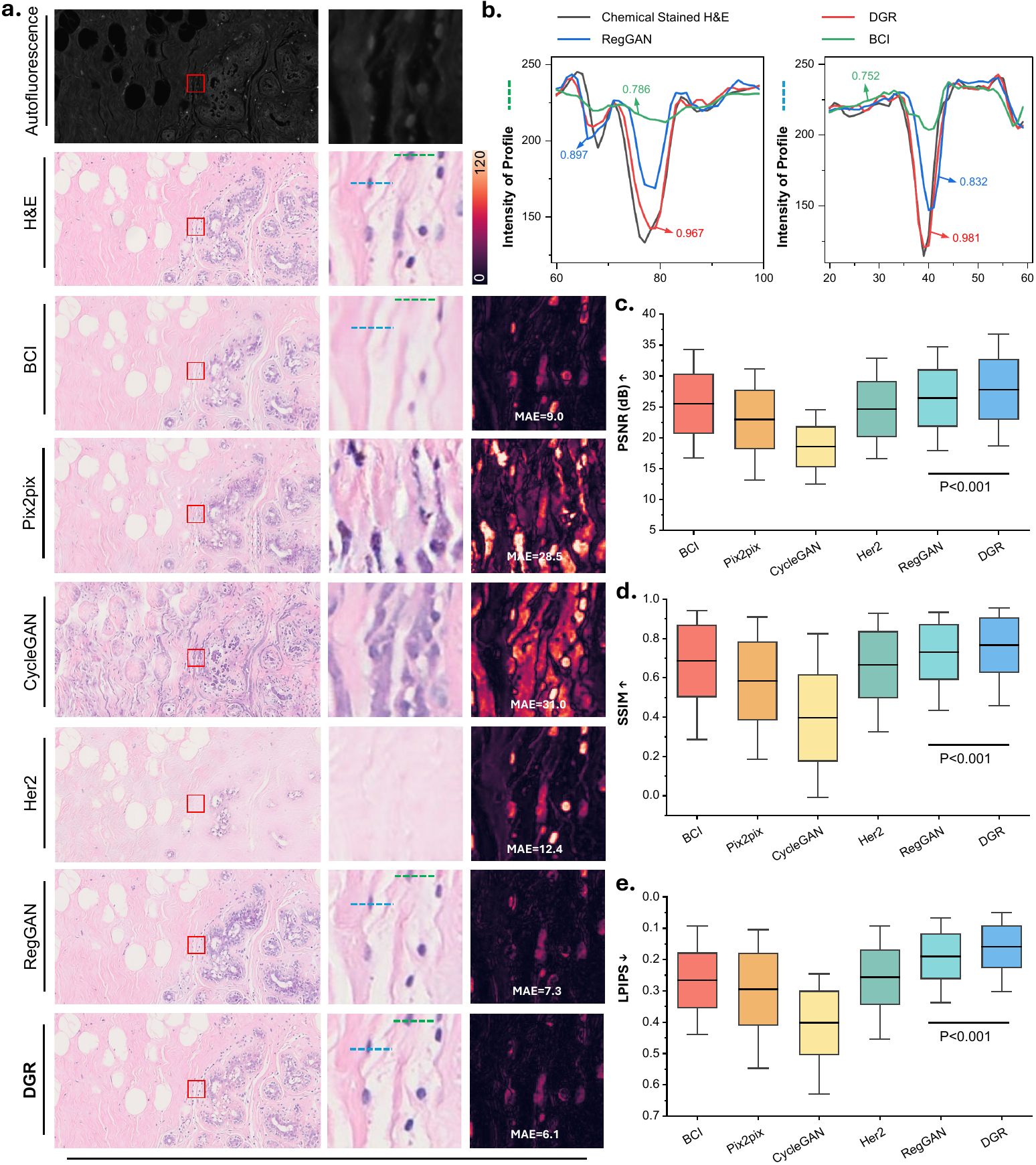}
\caption{\textbf{Comparison of different methods for virtual staining of AF images to H\&E images.} 
\textbf{a.} The original AF image, chemically stained H\&E image (GT), and virtually stained images generated by various methods. A heatmap of the mean absolute error (MAE) between the GT and generated images is also shown. Lower MAE indicates better performance. 
\textbf{b.} Intensity profiles along the dashed green and blue lines for the ground truth image and the top three performing models. The Pearson correlation coefficient (PCC) is provided for each method. 
\textbf{c-e.} Average performance of the methods evaluated using LPIPS, SSIM, and PSNR metrics, respectively.
The box limits represent the standard deviation, and the error bars indicate the 2.5\% and 97.5\% percentiles.
}
    \label{fig:af2he}
\end{figure*}

In this section, we evaluated the performance of our proposed method, DGR, across four internal datasets and one external dataset. 
To demonstrate its effectiveness, we compare DGR against the following four types of approaches: (1) the well-aligned pair-based method by Isola et al. \cite{isola2017image}, (2) the unpaired image method by Zhu et al. \cite{zhu2017unpaired}, (3) virtual staining models by Bai et al. \cite{bai2022label} and Liu et al. \cite{Liu_2022_CVPR}, and (4) the misalignment-robust model by Kong et al. \cite{kong2021breaking}.
DGR demonstrated consistently superior average performance on the five datasets (Fig. \ref{fig1}b) across all three metrics.
Specifically, it achieved a PSNR of 22.914 dB (+4.4\%, Fig. \ref{fig1}c), a structural similarity index (SSIM) of 0.676 (+4.0\%, Fig. \ref{fig1}d), and a LPIPS of 0.180 (+7.7\%, Fig. \ref{fig1}e).
Furthermore, we assessed the robustness of DGR under varying degrees of misalignment. 
The results show that DGR consistently outperformed the baseline across multiple datasets and various noise levels. 
Notably, under conditions of significant misalignment (±5° rotation, ±10\% translation, and ±10\% rescaling), DGR achieved a PSNR improvement of 3.4 dB (23.8\%) over the baseline (Fig. \ref{fig1}f).

\subsection{Staining Label-free Autofluorescence to H\&E }
In this task, we use autofluorescence (AF) images as input and virtually stain them to generate H\&E images.
From a quantitative perspective, the unsupervised method proposed by Zhu et al. \cite{zhu2017unpaired} demonstrates inferior performance due to its inability to ensure fidelity between the generated images and the ground truth (GT) images (Fig. \ref{fig:af2he}c-e, Extended Data Table \ref{tab:af2he}).
In contrast, traditional supervised approaches, such as the one by Isola et al. \cite{isola2017image}, outperform unsupervised models, emphasizing the advantages of supervised training. 
The virtual staining method proposed by Liu et al. \cite{Liu_2022_CVPR} using supervised training further surpasses universal image-to-image translation networks, achieving significant improvements.
However, for mentioned methods, roughly paired data can lead to inaccuracies when measuring the similarity between the generated images and the corresponding GT images. 
To address the issue of misalignment, Bai et al. \cite{bai2022label} introduced a multi-scale reconstruction loss, which improved performance. 
Similarly, Kong et al. \cite{kong2021breaking} incorporated a registration module to better align the generated images, significantly reducing misalignment and further enhancing performance.
Our proposed method, DGR, decouples the generation (virtual staining) and registration modules, allowing the generator to focus solely on staining. This design achieves superior performance across all three evaluation metrics (Fig. \ref{fig:af2he}). Notably, DGR's PSNR surpasses the next-best model (Kong et al. \cite{kong2021breaking}) by 1.4 dB (+5.3\%), demonstrating its effectiveness (Extended Data Table \ref{tab:af2he}).

To further evaluate the visual quality of the generated images, we compared virtually stained images with chemically stained images (Fig. \ref{fig:af2he}a). 
Images generated by Zhu et al. \cite{zhu2017unpaired}, Isola et al. \cite{isola2017image}, and Bai et al. \cite{bai2022label} exhibit notable discrepancies when compared to the GT images. 
While the methods by Liu et al. \cite{Liu_2022_CVPR} and Kong et al. \cite{kong2021breaking} achieve significantly lower mean absolute error (MAE) than the aforementioned approaches, their results suffer from a loss of detail. 
In contrast, images virtually stained by DGR show a closer resemblance to chemically stained H\&E images. 
Upon closer inspection, the images generated by DGR demonstrate superior alignment with the GT images.
Additionally, we analyzed the intensity profiles of the generated images produced by the top three best-performing models alongside the GT images (Fig. \ref{fig:af2he}b). 
The intensity values of DGR-generated images align more closely with those of the GT images, further underscoring DGR's effectiveness in restoring fine details in virtual staining.
Furthermore, the Pearson correlation coefficient (PCC) confirms that DGR's intensity profile is more consistent with the GT, reinforcing its superior performance.
\begin{figure*}[htbp]
    \centering
    \includegraphics[width=\textwidth]{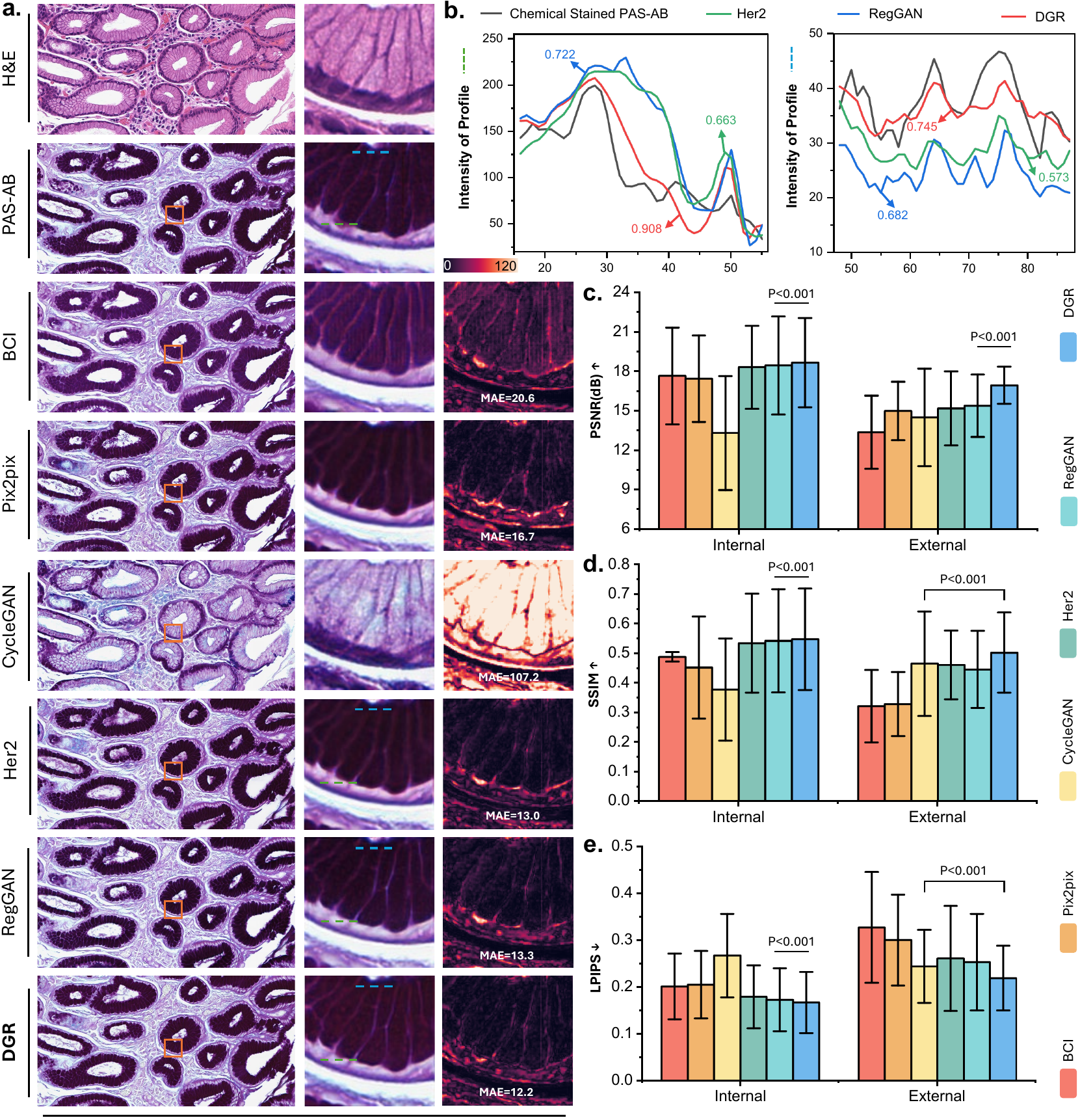}
\caption{\textbf{Results of various models for translating H\&E-stained images to PAS-AB-stained images.} 
\textbf{a}. The H\&E-stained image, chemically stained PAS-AB (GT) image, and virtually stained images generated by various methods.  
\textbf{b}. Intensity profiles along the dashed green and blue lines for the GT image and the top three performing models, along with the PCC values for these models.  
\textbf{c-e}. The average performance of various methods on both the internal and external test sets. The error bars indicate standard deviation.
}
    \label{fig:he2pas}
\end{figure*}

\subsection{Translating H\&E to PAS-AB}
In this task, we further evaluated the performance of various methods in converting H\&E images into Periodic Acid-Schiff-Alcian Blue (PAS-AB) images. 
Similar to the task of staining autofluorescence images into H\&E, Zhu et al.'s \cite{zhu2017unpaired} method demonstrated the poorest performance compared to other supervised approaches.
Among pixel-level supervised methods, our proposed method, DGR, consistently outperformed others across all three evaluation metrics (Fig. \ref{fig:he2pas}c-e, Extended Data Table \ref{tab:he2pas-ab}) in both internal and external test sets.
Specifically, DGR achieved a PSNR of 18.677 dB on the internal test set, surpassing Kong et al. \cite{kong2021breaking} by approximately 0.2 dB (+1.2\%). 
On the external test set, DGR achieved a significant improvement of approximately 1.6 dB (+10.1\%) compared to the second-best-performing model, Kong et al. \cite{kong2021breaking} (Extended Data Table \ref{tab:he2pas-ab}).

To further examine the performance of different methods, we visualized the virtually stained images. 
It is evident that Zhu et al.'s \cite{zhu2017unpaired} method struggles to accurately reproduce cellular components in the generated images, leading to significant discrepancies compared to real PAS-AB images (Fig. \ref{fig:he2pas}a). 
Additionally,  intensity of two profiles are provided (Fig. \ref{fig:he2pas}b). 
The results show that DGR's intensity values align most closely with the GT, achieving the highest PCC of 0.908 and 0.745, respectively.
Based on both quantitative and qualitative analyses, our findings demonstrate that the proposed DGR method is highly effective for virtually staining H\&E images into PAS-AB images.
\begin{figure*}[htbp]
    \centering
    \includegraphics[width=\textwidth]{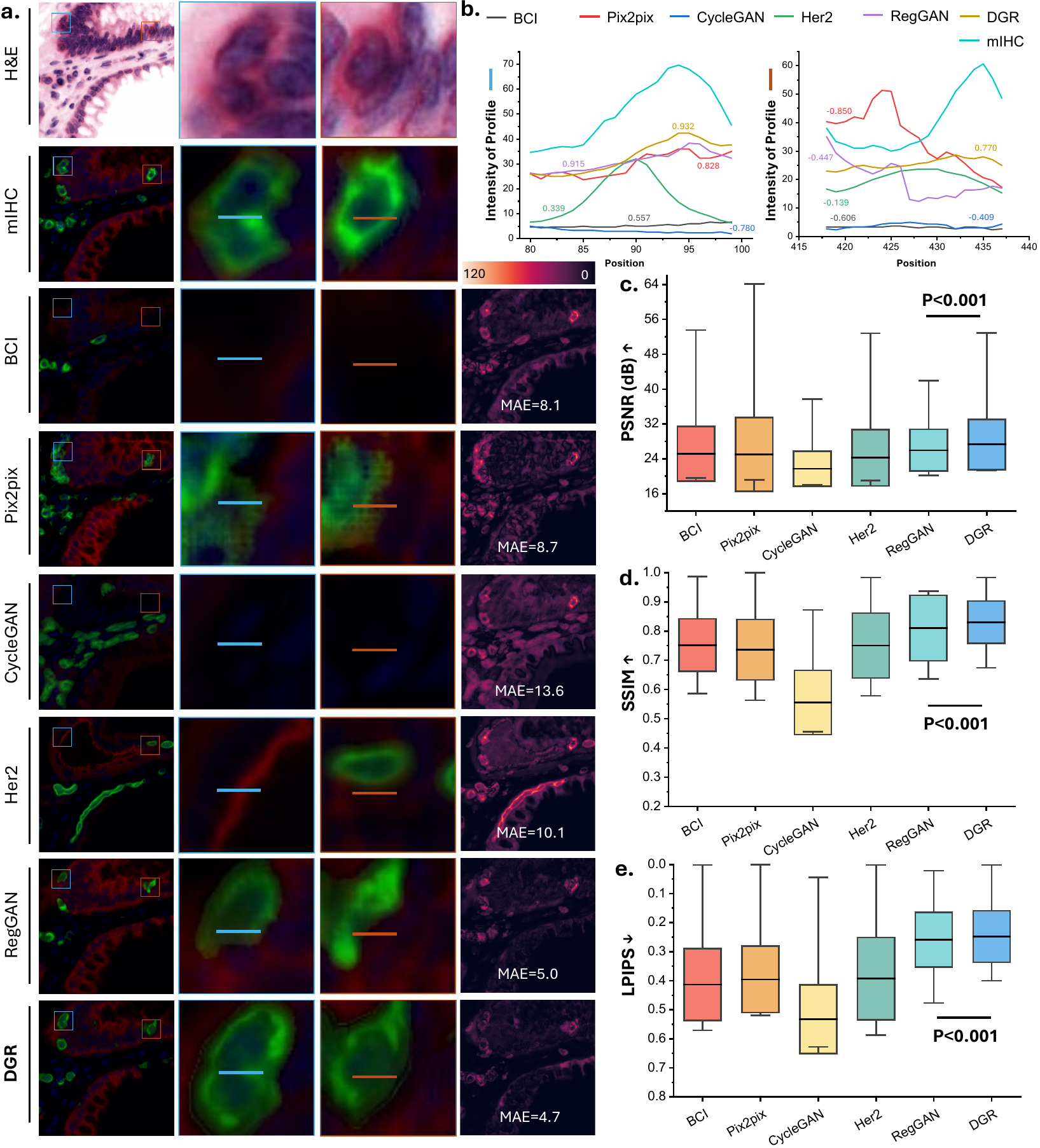}
     \caption{\textbf{Results of converting H\&E to mIHC images.}
    \textbf{a}. H\&E, mIHC, and corresponding virtually stained images predicted by various methods. The last column displays the mean absolute error (MAE) heatmap between the chemical stained mIHC image and the virtually stained mIHC image.
    \textbf{b}. Intensity profiles and PCCs for various methods.
    \textbf{c-e.} Average performance of different methods evaluated using PSNR, SSIM, and LPIPS metrics, respectively. The box limits represent the standard deviation, and the error bars indicate the 2.5\% and 97.5\% percentiles.
     }
    \label{fig:hemit}
\end{figure*}
 \subsection{Converting H\&E to mIHC}
Multiplex immunohistochemistry (mIHC) is a cutting-edge technique widely used in biomedical research and clinical diagnostics to simultaneously analyze multiple proteins within tissue samples.
This method significantly enhances our understanding of complex cellular interactions and the tumor microenvironment by enabling the visualization and quantification of various biomarkers in a single tissue section.
Translating H\&E stained images into multiplex mIHC images is an impressive yet highly challenging task. 
In this study, we leveraged a publicly available dataset to evaluate the potential of DGR in converting H\&E images into mIHC images.

Our statistical analysis shows that DGR achieves the highest PSNR, outperforming the second-best model, Kong et al., by approximately 1.3 dB (+5.1\%, Fig. \ref{fig:hemit}c). 
In addition to PSNR, DGR also demonstrates superior performance across two other key metrics (Fig. \ref{fig:hemit}d-e). These results highlight the critical role of image registration in improving virtual staining accuracy for roughly aligned image pairs.
Despite the higher PSNR values achieved by various methods on this dataset—compared to other datasets—this task remains challenging. 
For instance, Fig. \ref{fig:hemit}a illustrates images generated by different methods, revealing that only Kong et al. and DGR produce outputs that are relatively comparable to chemically stained mIHC images.
The intensity figures of two profiles also show that DGR is more consistent with GT (Fig. \ref{fig:hemit}b).
Specifically, DGR generates more reliable Pan Cytokeratin (panCK, represented in the red channel) images than Kong et al., while also excelling in 4',6-diamidino-2-phenylindole (DAPI, represented in the blue channel). 
When considering all three channels collectively, DGR demonstrates the best overall performance, underscoring its effectiveness in translating H\&E images into mIHC images.
 \begin{figure*}[htbp]
    \centering
    \includegraphics[width=\textwidth]{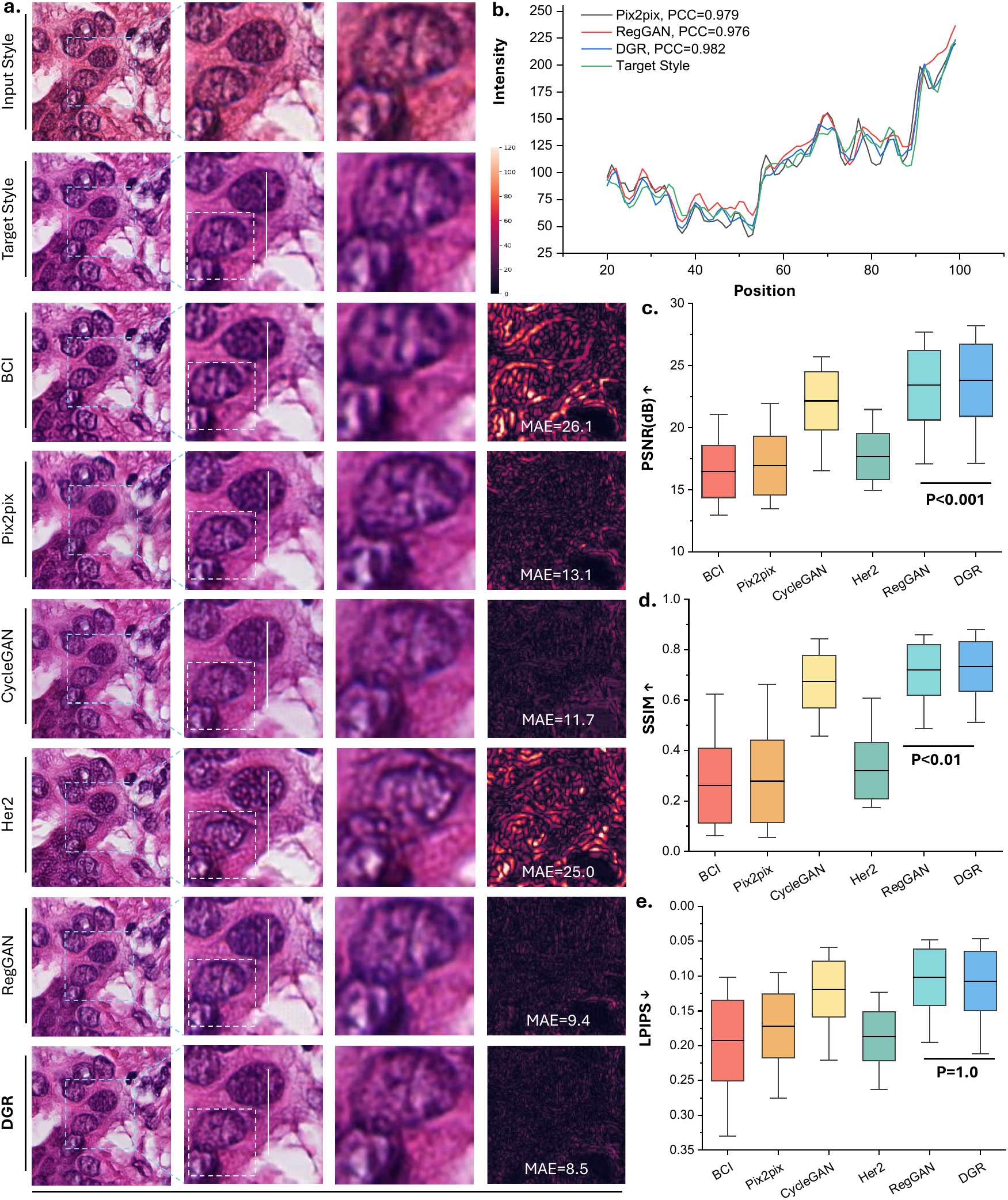}
    \caption{\textbf{Results of Normalizing Staining Styles.}
    \textbf{a}. Input style, target style, and corresponding normalized images generated by various methods. The last column displays the mean absolute error (MAE) heatmap between the target style image and the normalized image.
    \textbf{b}. Intensity profiles of the top three best-performing models. The PCC is calculated for these methods.
    \textbf{c-e.} Average performance of different methods evaluated using PSNR, SSIM, and LPIPS metrics, respectively. The box limits represent the standard deviation, and the error bars indicate the 2.5\% and 97.5\% percentiles.
    }
    \label{fig:aperio}
\end{figure*}
\subsection{Normalizing Staining Style of H\&E Image}
To further evaluate the effectiveness of DGR in other stain-related tasks, we conducted experiments using a stain normalization dataset \cite{kang2021stainnet}. 
The results show that DGR outperforms competing models in both PSNR and SSIM metrics, achieving approximately a 0.4 dB improvement (+1.7\%) in PSNR compared to the next best-performing model, Kong et al. (Fig. \ref{fig:aperio}c, d; Extended Data Table \ref{tab:aperio}).
In terms of perceptual similarity, DGR performs comparably to Kong et al., with only a marginal difference of 0.005 in Learned Perceptual Image Patch Similarity (LPIPS) metric (Fig. \ref{fig:aperio}e). 
Visual comparisons of normalized images generated by various methods are provided in Fig. \ref{fig:aperio}a. 
While this task is relatively less challenging than others, the results demonstrate that images normalized by Zhu et al., Kong et al., and DGR are significantly superior to those produced by other methods.
Among these, DGR achieves the lowest MAE of 8.5 (Fig. \ref{fig:aperio}c).
Furthermore, the intensity profile analysis (Fig. \ref{fig:aperio}b) reveals that DGR aligns most closely with the GT, achieving the highest PCC of 0.982. 
Both quantitative and qualitative results confirm that DGR consistently delivers superior performance, underscoring the effectiveness of the proposed method.
\subsection{Blinded Human Evaluation}
Both qualitative and quantitative results confirm that DGR can generate high-quality images compared to existing methods.
However, it cannot be concluded that the images generated by DGR are indistinguishable to pathologists.
To verify whether the generated images can deceive the eyes of pathologists, we randomly selected 250 virtually stained H\&E images and corresponding 250 chemically stained H\&E images, as well as 250 virtually stained PAS-AB images and corresponding 250 chemically stained PAS-AB images.
For each task, we randomly shuffled the 500 virtually stained and chemically stained images and asked an experienced pathologist to determine whether the images were real chemically stained ones.
Taking the H\&E images as an example, the confusion matrix (Fig. \ref{fig1}g) shows that it is difficult for the pathologist to distinguish between chemically and virtually stained H\&E images, with an accuracy of only 52\%.
We also conducted a Chi-Squared Test, which revealed no significant difference between the chemically and virtually stained images.
Similarly, in the PAS-AB task, the pathologist was unable to reliably distinguish between chemically and virtually stained images (Fig. \ref{fig1}h).
More generated H\&E and PAS-AB images are provided in Extended Data Fig. \ref{fig:af2hewsi} and Fig. \ref{fig:he2paswsi}, respectively.
In summary, DGR not only outperforms existing methods but also generates images that cannot be distinguished by experienced pathologists. 
This demonstrates the potential of virtual staining for real-world applications.
\subsection{Misalignment Resilience}
\begin{figure*}[htbp]
    \centering
    \includegraphics[width=\textwidth]{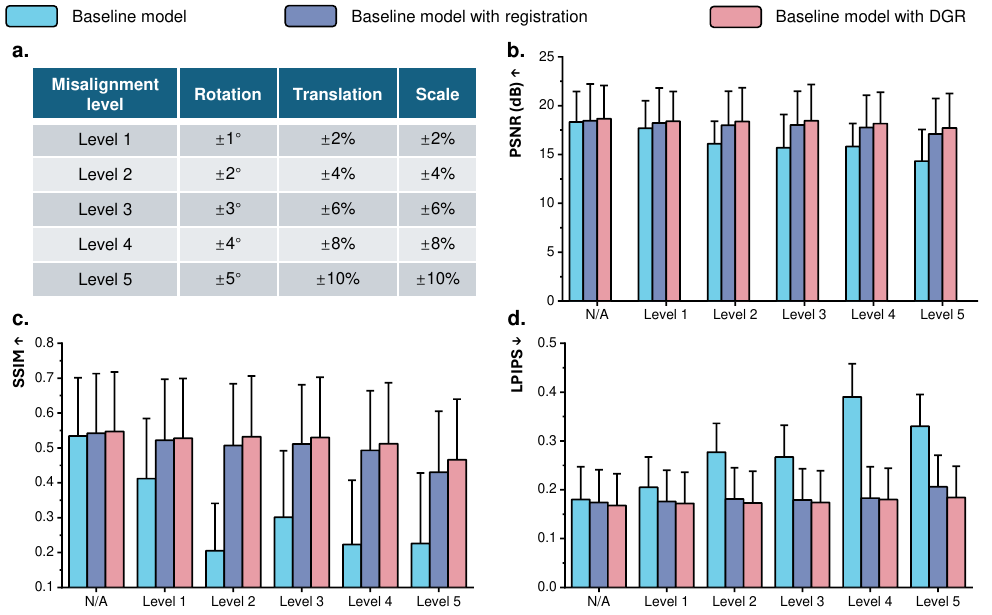}
    \caption{\textbf{Misalignment resilience of DGR across varying levels of misalignment.}
    \textbf{a.} The misalignment levels of experiments. 
    \textbf{b-d.} Overall results of the different strategies.
    The ``Baseline model'' contains no registration components. . 
    The ``Baseline model with registration'' incorporates the RnR module (similar to RegGAN).  
    The ``Baseline model with DGR'' includes both RnR and PcG modules for decoupled generation and registration. 
Our DGR framework demonstrates significantly improved robustness to misalignment, highlighting the importance of explicit spatial consistency constraints through the PcG module.
Detailed misalignment parameters and quantitative results are provided in Table\ref{ablation_study}. 
}
    \label{fig:noise_fig}
\end{figure*}
Pixel-level misalignment is a very common issue in real-world virtual staining applications. 
As a result, the practical utility of the DGR framework requires further validation across varying degrees of misalignment. 
To assess the robustness of our proposed method, we curated a well-aligned dataset comprising 11,918 paired H\&E and PAS-AB patches, each measuring 256 × 256 pixels. 
We then designed five levels of misalignment, incorporating rotations, translations, and scalings applied to the input images (Fig. \ref{fig:noise_fig}a). 
For model training, both input and target images were center-cropped to 128 × 128 pixels.

In our initial evaluation, we analyzed the baseline model \cite{bai2022label} and observed a consistent decline in performance across all three metrics as misalignment levels increased (Fig. \ref{fig:noise_fig}b-d, Extended Data Table \ref{ablation_study}). This trend underscores the baseline model’s lack of robustness when confronted with misaligned data. To address this, we integrated a registration module into the baseline model, inspired by the approach of Kong et al. \cite{kong2021breaking}. The results showed a reduced performance drop under significant misalignment compared to the baseline, indicating that the registration module mitigates the effects of misalignment (Fig. \ref{fig:noise_fig}b-d).
Building on this, we further enhanced the model by introducing a decoupling mechanism alongside the registration module. 
This addition led to improved performance on misaligned data. Notably, the Peak Signal-to-Noise Ratio (PSNR) exceeded 18 dB from misalignment Level 1 to Level 4, outperforming both the baseline model and the model with only the registration module. These findings highlight the effectiveness of the proposed decoupling strategy in bolstering the robustness of virtual staining for misaligned data, thereby advancing its applicability in real-world scenarios.
The effectiveness of the disentangled process is further visualized through the deformation fields generated by $R_1$ and $R_2$ (Extended Data Fig. \ref{fig:deformation_field}).
In conclusion, the DGR framework effectively disentangles generation and registration processes, yielding a more robust virtual staining model capable of handling misaligned image pairs. This enhancement marks a significant step forward in the development of reliable virtual staining techniques for practical applications.
\section{Discussion}
The utilization of virtual staining in pathology images eliminates the need for chemical reagents, offering a faster and more versatile way to generate multiple stains from an input image.
Recent deep learning models, such as U-Net \cite{ronneberger2015u} and its variants, has significantly improved the performance of virtual staining \cite{bai2022label, liu2020global, Liu_2022_CVPR,rivenson2019virtual,christiansen2018silico, ounkomol2018label}. 
These models have been successfully applied to various virtual staining tasks \cite{bai2023deep,zhang2020digital, christiansen2018silico,ma2023efficient}. 
However, these methods are subject to certain limitations, including the need for well-aligned pairs of images for model training, which may not always be readily available.
In contrast to Pix2Pix-like methods, cycle-consistent adversarial frameworks \cite{liu2017unsupervised, huang2018multimodal, zhu2017unpaired} can be used to train the model without paired images.
For example, Liu et al. \cite{liu2021generation} adopted an improved Cycle-GAN to translate H\&E images to immunohistochemical(IHC) images. 
Nevertheless, it is important to acknowledge that the cycle-consistent loss places a greater emphasis on achieving global style coherence in generated images, which can not guarantee high-fidelity reproduction of the original content, thus potentially limiting its utility in misaligned images.
Therefore, it remains a significant challenge to train a virtual staining model using misaligned paired images.

Recently, image registration techniques \cite{kim2019unsupervised, balakrishnan2019voxelmorph, hering2022learn2reg, chen2022transmorph, shen2019networks, mok2020fast} have been integrated into virtual staining to enhance performance by aligning generated images with target images \cite{kong2021breaking, menze2014multimodal}. However, naive registration approaches, lacking constraints on the input and generated images, do not ensure proper alignment. To address this issue, we introduced a framework called DGR that decouples the generation and registration processes in current virtual staining methods for pathology images. We evaluated the feasibility and performance of our proposed method against other virtual staining techniques using five datasets. 
The experimental results demonstrate the effectiveness of DGR in datasets with varying degrees of misalignment. Our method shows promise for virtual staining of misaligned paired images.

However, the generalization capability of the evaluated method still has room for improvement when applied to external datasets. 
Additionally, the current DGR framework requires relatively higher computational resources compared to some baseline methods, which may limit its deployment in resource-constrained scenarios (Extended Data Table \ref{tab:efficiency}).
To address these limitations, future research should focus on curating large-scale, diverse datasets to enhance robustness and adaptability, along with optimizing the model architecture to improve efficiency. 
Developing advanced generative AI models, such as diffusion models, for virtual staining also presents a promising opportunity to standardize the virtual staining process across various stain types. 
By expanding dataset size, leveraging cutting-edge generative AI techniques, and improving computational efficiency, we can produce virtually stained images that rival the quality of chemically stained ones, while avoiding the inherent drawbacks of chemical staining.

\section{Method}
\subsection{Problem Setting}
To aid in the illustration of our method, we introduce the relevant notation in this section. 
A misaligned image pair is denoted as $(x_n, y_n)$, where $x_n$ and $y_n$ belong to two different modalities and are spatially misaligned.
We assume the existence of a well-aligned counterpart $\hat{y}_n$ for each $x_n$; however, $\hat{y}_n$ is not available in real-world scenarios. 
Our objective is to train a model using the misaligned dataset $[(x_1, y_1), ..., (x_N, y_N)]$, which consists of $N$ pairs, and achieve performance comparable to that of the well-aligned dataset $[(x_1, \hat{y}_1), ..., (x_N, \hat{y}_N)]$.
\subsection{Baseline}
The virtual staining methods based on the Pix2Pix framework \cite{isola2017image} rely on pixel-level $L_1$ loss for optimization, as shown in Equation \ref{eq_p2p}. 
However, this framework does not account for misalignment that may be present in the paired images. 
Therefore, the performance of the framework is heavily dependent on the degree of alignment between the paired images.
\begin{equation}
    \label{eq_p2p}
    \min_{G} \mathcal{L}(x, y) = \frac{1}{N}\sum_{n=1}^N ||G(x_n)- y_n||_1  
\end{equation}
To alleviate the influence of misalignment, Kong et al.\cite{kong2021breaking, song2022learning} proposed RegGAN based on the idea of ``loss correction"\cite{patrini2017making} for the magnetic resonance  image translation task.
This work introduced a registration network ($R$) to estimate the deformation field $\phi$ between the generated image $G(x)$ and the noisy ground truth $y$, and then use $\phi$ to re-sample $G(x)$ for loss computation. This process can be represented by the following formula:

\begin{equation}
    \phi_n = R(G(x_n), y_n)
\end{equation}
\begin{equation}
    \label{eq_RegGAN_m}
    \min_{G, R} \mathcal{L}(x, y) = \frac{1}{N}\sum_{n=1}^N ||\phi_n \circ G(x_n) - y_n||_1,
\end{equation}
where $\circ$ represents a re-sampling operator.
In the RegGAN, the authors assumed the noisy ground truth can be expressed as a displacement error, i.e., $y=\phi \circ \hat{y}$.
However, it is worth noting that Equation \ref{eq_RegGAN_m} demonstrates that the RegGAN method only guarantees the re-sampled generated image $\phi \circ G(x)$ is well-aligned with the noisy ground truth $y$, and does not explicitly ensure that $G(x)$ is well-aligned with the corresponding ideal ground truth $\hat{y}$.
Since $\hat{y}$ is well aligned with the input image $x$, the misalignment between $G(x)$ and $\hat{y}$ also implies that the generated image may be not well aligned with the input image.
Figure \ref{fig:DTR-arch}a illustrates the limitations of solely relying on the registration module (Kong et al. \cite{kong2021breaking}) and presents the ideal solution. 
To address this limitation, we propose a novel approach that incorporates additional constraints to ensure the alignment between $G(x)$ and ideal ground truth $\hat{y}$.
\subsection{Decoupling Generation and Registration}
Based on the analysis of RegGAN, our findings suggest that the generation and registration processes are inherently interdependent. 
This occurs because the correction loss function is applied to the re-sampled image $\phi \circ G(x)$, while the generated image $G(x)$ itself is not subject to strong constraints such as a correction loss.
As a result, the generation and registration stages must be carefully balanced and coordinated to achieve optimal results.
Drawing inspiration from GAN models \cite{goodfellow2020generative}, we introduce an adversarial registration process to decouple generation and registration ensuring that the generated image $G(x)$ is accurately aligned with the ideal output $\hat{y}$. 
The framework of our method is shown in Extended Data Fig. \ref{fig:DTR-arch}b including registration for noise reduction module and position-consistency generation module for forcing generator to stain input image only without changing the position of pixels.

\textbf{Registration for Noise Reduction.} 
To reduce the influence of misalignment between $G(x)$ and $y$, we first adopt a registration network ($R_1$) to align the generated image $G(x)$ with the ``noisy" ground truth $y$.
With the assistance of registration for noise reduction, the reconstruction loss can be represented as following equation:
\begin{equation}
    \label{eq_R1}
   \mathcal{L}_{L1}(x, y) = \frac{1}{N}\sum_{n=1}^N||R_1(G(x_n), y_n) \circ G(x_n) - y_n||_1,
\end{equation}
where $R_{1}(G(x_n), y_n$) is the deformation field between generated image $G(x)$ and ``noisy" ground truth $y$. 
Additionally, the structural similarity (SSIM) metric is also adopted to improve the structural similarity between generated images and target images.
To use this metric, we utilize following equation:
\begin{equation}
    \begin{split}
    \mathcal{L}_{ssim}(x, y) = -\frac{1}{N}\sum_{n=1}^{N}log(1+\\ssim(R_1(G(x_n), y_n)\circ G(x_n), y_n))
    \end{split}
\end{equation}
where \textit{ssim} is function that can calculate the structural similarity between two images. 

To prevent overfitting of the deformation field predicted by the registration network $R_1$, we utilize a smoothness regularization as shown in the following equation:
\begin{equation}
    \label{eq_smooth_r1}
    \mathcal{L}_{smooth}(x, y) = \frac{1}{N}\sum_{n=1}^N||\nabla R_{1}(G(x_n), y_n)||_2, 
\end{equation}
where $\nabla$ indicates differential operator.
To make the model generate realistic images, we also adopt generative adversarial training, the adversarial loss for the generator $G$ and the discriminator $D$ is shown in Equations \ref{eq_adv} and \ref{eq_d}.
\begin{equation}
    \label{eq_adv}
    \mathcal{L}_{adv}(x) = -\frac{1}{N}\sum_{n=1}^N log(D(G(x_n)))
\end{equation}
\begin{equation}
    \label{eq_d}
    \mathcal{L}_{dis}(x, y) = -\frac{1}{N}\sum_{n=1}^N [log(1 - D(G(x_n)) + log(D(y_n))]
\end{equation}
The total loss for the Registration for Noise Reduction (RNR) is expressed as the following equation.
\begin{equation}
    \label{eq_ra}
    \begin{split}
    \mathcal{L}_{RNR}(x, y) = \alpha\mathcal{L}_{L1}(x, y) + \beta\mathcal{L}_{ssim} (x, y) + \\\gamma\mathcal{L}_{smooth}(x, y) + \delta\mathcal{L}_{adv}(x)
    \end{split}
\end{equation}

\textbf{Position-consistency Generation.} 
With the constraint imposed by the registration for noise reduction module, the re-sampled generated image $\phi \circ G(x)$ and the ``noisy" ground truth $y$ are encouraged to be aligned, which improves the credibility of the reconstruction loss (e.g., $\mathcal{L}_{L1}$).
However, a critical limitation persists: there is no explicit constraint ensuring that the \textit{generated image itself}, $G(x)$, is aligned with the \textit{input image}, $x$.
This absence of direct constraint can lead to a coupled optimization where the generator $G$ learns to ``hide" structural misalignment in its output, expecting the subsequent registration network $R_1$ to correct it.
This coupling contradicts the desired objective of a virtual staining model, which should perform a pure color transformation \textit{without} altering the spatial position of cellular structures.
The intuition behind our approach is that an ideal virtual staining process should be \textit{spatially consistent}. Since the input image $x$ (e.g., autofluorescence) and the target stain $y$ (e.g., H\&E) are acquired from the exact same physical tissue section, they are inherently paired and share an identical underlying biological structure.
The primary difference between them is their visual appearance (modality), not their geometry.
Therefore, a correctly generated image $G(x)$ must constitute a pixel-wise translation of the input $x$ into the target stain's style.
Any spatial misalignment between $G(x)$ and $x$ is not a property of the data but an artifact of an imperfectly trained generator.
Our goal is to enforce this spatial consistency, ensuring the generator learns to change \textit{only} the stain appearance, analogous to applying a color filter that does not distort the image.

To achieve this goal, we design a constraint as defined in Equation \ref{eq_r2_g} to optimize $R_2$.
\begin{equation}
    \label{eq_r2_g}
    \mathcal{L}_{T1}(x) = \frac{1}{N}\sum_{n=1}^N ||R_{2}(x_n, G(x_n))\circ G(x_n) - G(x_n)||_1
\end{equation}
However, relying solely on this constraint will prevent the registration network $R_2$ from learning informative deformation field. 
$R_2$  could simply produce an identity deformation field for any given paired images, failing to extract useful information. 
To prevent $R_2$ from circumventing the learning process, we need an additional constraint that compels $R_2$ to accurately predict the true deformation field.

From Equation \ref{eq_r2_g}, we can see that generated image $G(x)$ and the input image $x$ will be well aligned.
This alignment  allows us to conclude that the deformation field between input image $x$  and target image $y$  is identical to the deformation field between generated image $G(x)$ and target image $y$. 
This property can be leveraged to facilitate adversarial training of the registration network $R_2$.
To this end, we utilize the loss function defined in Equation \ref{eq_ti_diff}:
\begin{equation}
    \begin{split}
        \label{eq_ti_diff}
    \mathcal{L}_{T2}(x, y) = \frac{1}{N}\sum_{n=1}^N ||R_{2}(x_n, y_n)\circ G(x_n) \\ - R_{1}(G(x_n), y_n)\circ G(x_n)||_1       
    \end{split}
\end{equation}
Here, we aim to minimize the $L_1$ distance between the deformation field of $x_n$ and $y_n$, and the deformation field of $G(x_n)$ and $y_n$. 
This loss function encourages registration network $R_2$ to effectively learn how to align cross-modal images (i.e., the input image and its corresponding target image).  
Additionally, similar to Equation \ref{eq_smooth_r1}, we also adopt smooth regularization for $R_2$ to avoid overfitting.
\begin{equation}
    \label{reg_21}
        \mathcal{L}_{T1R}(x) = \frac{1}{N}\sum_{n=1}^N||\nabla R_{2}(x_n, G(x_n))||_2 
\end{equation}
\begin{equation}
    \label{reg_22}
        \mathcal{L}_{T2R}(x, y) = \frac{1}{N}\sum_{n=1}^N||\nabla R_{2}(x_n, y_n)||_2
\end{equation}

\textbf{Training Strategy.}
The above registration for noise reduction and position-consistency generation modules include generator $G$, discriminator $D$, registration networks $R_1$ and $R_2$.
We regard training as two adversarial processes.
Firstly, we update the parameters of generator $G$ and registration network $R_1$, the loss can be represented as:
\begin{equation}
    \min_{G, R_1} \mathcal{L}(x, y) = \mathcal{L}_{RNR}(x, y) + \epsilon\mathcal{L}_{T1}(x).
\end{equation}
Based on this loss, we can infer that the re-sampled image $R_1(G(x_n), y_n) \circ G(x_n)$ will be aligned well with the target image $y$, however, it can not ensure that the input image $x$ can be well aligned with the translated image $G(x)$ due to that $R_2$ could simply output identity mapping for all images to minimize $\mathcal{L}_{T1}(x)$.
To avoid laziness of $R_2$,  we update the registration network $R_2$ using constraints described in the position-consistency generation module as shown in following equation.
\begin{equation}
    \begin{split}
    \min_{R_2}\mathcal{L}(x, y) = \mathcal{L}_{T1}(x) + \gamma\mathcal{L}_{T1R}(x)+\\\mathcal{L}_{T2}(x, y)+\gamma\mathcal{L}_{T2R}(x, y)    
    \end{split}
\end{equation}
Finally, we update the discriminator using the loss function defined in Equation \ref{eq_d}.

This training strategy enables us to jointly train the generator and both registration networks, while also ensuring that the generated image is well-aligned with the input image. 
By incorporating adversarial components into our training process, our model can effectively learn informative features and achieve superior performance.
\subsection{Visualization of decoupling strategy}
To gain a clear understanding of proposed DGR model, we intentionally introduced misalignment into the paired data.
Then we visualized the generated images and the related registered images (Extended Data Fig. \ref{fig:DTR-arch}c). 
It is evident that the input image $x$ and the generated image $G(x)$ are properly aligned, meanwhile the registered image $G(x)_{M1}$ also shows alignment with the target image $y$.
The decrease of mean absolute error (MAE) of error map $|G(x)_{M1}-y|$ compared to $|G(x)-y|$ further supports the efficacy of registration network $R_1$.
To evaluate the effectiveness of $R_2$, the error maps $|G(x)-G(x)_{M2}|$ and $|G(x)_{M1}-G(x)_{M3}|$ are also visualized. 
The MAE between $G(x)$ and $G(x)_{M2}$ demonstrates that the deformation field $M_2$ produced by $R_2$ with input $x$ and $G(x)$ closely resembles an identity mapping, aligning with our intended objective. 
However, challenges persist for $R_2$ in achieving perfect alignment between input and target images, indicating potential for improvement in its performance.
\subsection{Datasets}
To validate the effectiveness of DGR, we adopted five datasets for various virtual staining tasks. 
The size and number of images are reported in Table \ref{tab:data_info}.\\\\
\textbf{AF2HE Dataset.} The AF2HE dataset comprises 15 whole slide images (WSIs) of breast and lung cancer tissue samples  \cite{dai2024exceeding}. 
The specimens were initially imaged in their native autofluorescence (AF) state using a widefield microscopy system equipped with a 265 nm ultraviolet excitation source and a 10× objective lens, enabling high-contrast nuclear visualization without staining \cite{zhang2022high}. 
Following AF imaging, the same slides were subjected to standard hematoxylin and eosin (H\&E) staining and rescanned under a 20× objective lens using a whole-slide scanner. The image pairs were rigorously aligned using the VALIS framework to ensure spatial correspondence \cite{gatenbee2023virtual}.
Subsequently, the registered WSIs were partitioned into 128 × 128 patches, resulting in a total of 50,447 paired patches for training and 4,422 for testing. 

\noindent
\textbf{HE2PAS-AB Dataset.} This dataset was collected from clinical specimens at the Prince of Wales Hospital in Hong Kong. 
It consists of WSIs from which a total of 10,727 aligned patch pairs (128 × 128 pixels) were extracted for training and 1,191 for testing.  
Each tissue section underwent sequential staining: initially with H\&E, followed by de-staining via acid alcohol treatment, and finally restaining with PAS-AB. 
All WSIs were aligned using the VALIS framework to ensure precise spatial registration between staining modalities. 
To rigorously assess model generalizability across different patient populations within the same institution, an external validation set containing 2,841 patches was additionally constructed. 
These patches were sourced from separate tissue sections and distinct patient cohorts that did not overlap temporally or spatially with the training or test samples, thereby ensuring independent evaluation. 

\noindent
\textbf{HEMIT Dataset.} 
The dataset employed in this work is derived from the collection introduced by Bian et al.  \cite{bian2024hemit}, consisting of histologically paired H\&E and multiplex immunohistochemistry (mIHC) whole-slide images originating from the ImmunoAlzer \cite{bian2021immunoaizer}. 
A key advantage of this dataset is that both staining protocols were applied sequentially on the same tissue section, thereby minimizing structural discrepancies and ensuring high pixel-level alignment—a notable improvement over conventional consecutive-section approaches. 
The mIHC staining incorporates three biologically relevant markers: DAPI for nuclear visualization, panCK for tumor region delineation, and CD3 for T-cell identification, each corresponding to a distinct channel in the image (blue, red, and green, respectively). 
We adopt the official partitioning of the data into training, validation, and test sets with 3717, 630, and 945 image pairs, respectively, ensuring that each partition contains tissue samples from different patients to prevent data leakage. 
To reduce computational demands during model training, we extract a central region of 512×512 pixels from each original 1024×1024 patch. 
Additionally, all H\&E images undergo stain normalization to enhance consistency and generalization across samples.

\noindent
\textbf{A2H Dataset.} The A2H dataset was re-organized by Kang et al. \cite{kang2021stainnet}, and the images were obtained from the MITOS\_ATYPIA ICPR-14 challenge \cite{roux2014detection}. 
Original WSIs were stained with H\&E and digitized using two distinct slide scanners (Aperio Scanscope XT and Hamamatsu Nanozoomer 2.0-HT).  
The processed images have a size of $256 \times 256$. 
This dataset consists of 19,200 pairs for training and 7,936 pairs for testing.
It is used to evaluate the performance of the proposed method in the stain normalization task.\\\\
\textbf{Well Aligned HE2PAS-AB Dataset}
To investigate the resilience of DGR on datasets with varying degrees of misalignment, we extracted paired patches of size 512$\times$512 from H\&E and PAS-AB WSIs. These patches were then registered, resulting in a training set of 10,727 pairs and a test set of 1,191 pairs, each sized 256$\times$256 pixels. Through this registration process, the images were aligned to the greatest extent possible.
Subsequently, we introduced five levels of misalignment involving rotations, translations, and scalings to the input images. The input and target images were then center-cropped to 128$\times$128 pixels for model training. Detailed configurations for these misalignments are provided in Extended Data Table \ref{ablation_study}.
\subsection{Experiments Setting}
For the models utilized in this study, we adhered to the hyperparameters recommended in the original papers for training. 
Our method employs the Adam optimizer \cite{kingma2014adam} for all models. Specifically, for the AF2HE, HE2PAS-AB, and A2H datasets, the learning rate for $G$ is set to 5e-5, while for the remaining  HEMIT dataset, the learning rate is decayed to 1e-5.
A comprehensive list of the hyperparameters is provided in Table \ref{tab:hyper}. 
All experiments were conducted using PyTorch \cite{paszke2019pytorch} on a workstation equipped with eight NVIDIA GeForce RTX 3090 GPUs. \\\\
\noindent

\noindent
\textbf{Metrics and Statistical Analysis } To evaluate the generated images, we adopt Peak Signal to Noise Ratio (PSNR), Structural Similarity (SSIM), and Learned Perceptual Image Patch Similarity (LPIPS) \cite{zhang2018perceptual}.
It is important to note that PSNR and SSIM primarily assess pixel-level similarity between two images, whereas LPIPS focuses on perceptual similarity. Additionally, we utilized a two-tailed independent samples t-test for statistical analysis unless other methods were explicitly stated. 
\section{Data availability}
In this study, we utilize five datasets. The HEMIT and A2H datasets consist of paired patches and can be downloaded from \href{https://github.com/BianChang/HEMIT-DATASET}{https://github.com/BianChang/HEMIT-DATASET} and \href{https://github.com/khtao/StainNet}{https://github.com/khtao/StainNet}, respectively. The AF2HE dataset is available at \href{https://github.com/TABLAB-HKUST/U-Frame}{https://github.com/TABLAB-HKUST/U-Frame}, although full access may require submission of an application. The HE2PAS-AB and HE2PAS-AB-EXT datasets were collected from the Prince of Wales Hospital in Hong Kong.
Due to privacy concerns, these datasets are not publicly available; however, interested readers may contact the corresponding author to request access.
\section{Code availability}
The code and model weights trained on different datasets are available on GitHub (\href{https://github.com/birkhoffkiki/DTR}{https://github.com/birkhoffkiki/DTR}) 

\section{Ethics declarations}
This project has been reviewed and approved by the Human and Artefacts Research Ethics Committee (HAREC) of Hong Kong University of Science and Technology. The protocol number is HREP-2024-0212.
\section{Author Contribution}
J.M. conceived the study, designed the experiments, and wrote the manuscript. 
W.L. implemented other state-of-the-art methods for comparison. 
J.L. and L.L. conducted the blinded human evaluation and provided medical guidance. 
Z.L. and L.W. contributed to the experimental design and reviewed the manuscript. 
F.Z. provided advice on visualization and figure preparation. R.C.K.C. supplied data and offered medical guidance. 
T.T.W.W. provided autofluorescence data. 
H.C. supervised the project.

\section{Acknowledgments}
This work was supported by National Natural Science Foundation of China (No. 62202403), Hong Kong Innovation and Technology Commission (Project No. PRP/034/22FX and ITCPD/17-9), and Frontier Technology Research for Joint Institutes with Industry Scheme (No. OKT24EG01).
\noindent

\begin{appendices}
\section{Extended Data}
\begin{table*}[hbp]
    \centering
     \caption{The details of adopted datasets for internal and external validation. The HE2PAS-AB-EXT is not used for external validation.}
    \begin{tabular}{llll}
    \hline
         Dataset&   Image Size& Train Number&Test Number\\
    \hline
         AF2HE& 
     $128\times 128$& 50,447&4,422\\
 HE2PAS-AB& $128\times 128$& 10,727&1,191\\
 HE2PAS-AB-EXT& $ 128\times 128$& N/A&2,841\\
 HEMIT& $512\times 512$& 3,717&945\\
 Aperio2Hamamats& $256\times 256$& 19,200&7,396\\
 \hline
 \end{tabular}
    \label{tab:data_info}
\end{table*}
\begin{table*}[hbp]
    \centering
    \caption{The hyperparameters used for training DGR. }
    \begin{tabular}{ll}
         \hline
         &  values\\
         \hline
         learning rate for $D$&  1e-5\\
         learning rate for $R_1$&  1e-5\\
         learning rate for $R_2$&  1e-5\\
         loss weight $\alpha$&  10\\
 loss weight $\beta$&0.5\\
 loss weight $\gamma$&10\\
 loss weight $\delta$&0.1\\
 loss weight $\epsilon$&5\\
 Epoch&200\\
 \hline
    \end{tabular}

    \label{tab:hyper}
\end{table*}
\begin{table*}[hbp]
\centering
\caption{Comparison of computational efficiency across different models (image size: 256×256).
GPU memory usage is measured with a batch size of 1 during both inference and training stages.
Inference speed is evaluated with a batch size of 32, averaged over 1000 runs to compute the processing time for 32 images.
Training speed is measured with a batch size of 1, averaged over 1000 runs to compute the processing time per image. 
The results are tested using a NVIDIA L20  GPU.
}
\label{tab:efficiency}
\begin{tabular}{lllll}
\hline
& Inference Memory& Inference Speed&  Training Memory&Training Speed\\
\hline
BCI \cite{Liu_2022_CVPR}& 104.73 MB &4.042 ms &448.14 MB&29.796 ms\\
Pix2pix \cite{isola2017image}&245.94 MB&0.501 ms&1051.48 MB&11.912 ms\\
CycleGAN \cite{zhu2017unpaired}&104.73 MB&4.042 ms&448.14 MB &29.926 ms\\
Her2 \cite{bai2022label}& 602.78 MB & 9.622 ms & 1403.90 MB & 51.007 ms\\
RegGAN \cite{kong2021breaking}& 108.73 MB &4.040 ms & 448.14 MB &28.870 ms\\
 UNSB \cite{kim2023unpaired}
&137.76 MB& 4.866 ms& 543.95 MB&30.216 ms\\
 StegoGAN \cite{wu2024stegogan}
&119.50 MB &4.479 ms &478.89 MB &32.321 ms\\
 ASP \cite{adaptive_Supervised_patch}
& 137.79 MB & 5.706 ms & 676.99 MB & 27.800 ms\\
 ST \cite{stain_transformation}& 43.69 MB& 0.731 ms& 135.49 MB& 31.446 ms\\
DGR& 602.78 MB & 9.616 ms & 1441.39 MB & 70.139 ms \\
\hline
\end{tabular}
\end{table*}

\begin{table*}[hbp]
\centering
\caption{Average performance of different methods on four internal test set.}
\label{tab:af2he}
\begin{tabular}{llllll}
\hline
& PSNR$\uparrow$ & SSIM$\uparrow$ & LPIPS$\downarrow$  & FID$\downarrow$&KID$\downarrow$\\
\hline
BCI \cite{Liu_2022_CVPR}&21.198&0.547&0.269 & 45.793&0.031\\
Pix2pix \cite{isola2017image}&20.599&0.513&0.267 & 32.805&0.019\\
CycleGAN \cite{zhu2017unpaired}&18.945&0.501&0.331 & 50.957&0.043\\
Her2 \cite{bai2022label}& 21.221& 0.568&0.254 & 45.891&0.031\\
RegGAN \cite{kong2021breaking}& 23.572& 0.701&0.181 & \pmb{25.839}&\pmb{0.012}\\
 UNSB \cite{kim2023unpaired}
& 18.894& 0.517&0.336 & 33.642&0.019\\
 StegoGAN \cite{wu2024stegogan}
& 19.373& 0.552&0.358 & 47.562&0.030\\
 ASP \cite{adaptive_Supervised_patch}
& 19.636& 0.550&0.279 & 36.055&0.022\\
 ST \cite{stain_transformation}& 21.322& 0.505&0.250 & 81.251&0.066\\
DGR&\pmb{24.402}&\pmb{0.719}&\pmb{0.171} & 28.593&0.015\\
\hline
\end{tabular}
\end{table*}

\begin{table*}[hb]
\centering
\caption{Performance of different methods on staining label-free autofluorescence images to H\&E images. The mean and 95\% CI are reported.}
\label{tab:af2he}
\begin{tabular}{llll}
\hline
& PSNR$\uparrow$ & SSIM$\uparrow$ & LPIPS$\downarrow$  \\
\hline
BCI \cite{Liu_2022_CVPR}&25.487 (25.345-25.617)&0.686 (0.680-0.691)&0.266 (0.264-0.269) \\
Pix2pix \cite{isola2017image}&22.962 (22.815-23.095)&0.584 (0.579-0.590)&0.295 (0.291-0.298) \\
CycleGAN \cite{zhu2017unpaired}&18.538 (18.442-18.630)&0.395 (0.389-0.402)&0.402  (0.399-0.405) \\
Her2 \cite{bai2022label}& 24.623 (24.495-24.753)& 0.667 (0.662-0.672)&0.257 (0.254-0.259) \\
 RegGAN \cite{kong2021breaking}& 26.422 (26.284-26.553)& 0.731  (0.727-0.735)& 0.189 (0.188-0.192) \\
UNSB \cite{kim2023unpaired}& 18.513 (18.404-18.619)& 0.452 (0.446-0.458)&0.433 (0.430-0.437)\\
 StegoGAN \cite{wu2024stegogan}& 21.867 (21.774-21.962)& 0.575 (0.570-0.579)& 0.493 (0.491-0.495)\\
 ASP \cite{adaptive_Supervised_patch}& 17.794 (17.694-17.894)&  0.402 (0.395-0.408)& 0.412 (0.408-0.415)\\
 ST \cite{stain_transformation}& 25.650 (25.512-25.793)& 0.655 (0.649-0.660)&  0.238 (0.236-0.240)\\
DGR&\pmb{27.803  (27.660-27.947)}&\pmb{0.766  (0.762-0.770)}&\pmb{0.159 (0.157-0.161)} \\
\hline
\end{tabular}
\end{table*}

\begin{table*}[hb]
\centering
\caption{Quantitative evaluation of different methods for translating autofluorescence images to H\&E. Performance is measured using the Fréchet Inception Distance (FID) and the Kernel Inception Distance (KID, reported as mean $\pm$ standard deviation).}
\label{tab:af2he}
\begin{tabular}{lll}
\hline
& FID$\downarrow$&KID$\downarrow$\\
\hline
BCI \cite{Liu_2022_CVPR}& 56.121&0.041$\pm$0.006\\
Pix2pix \cite{isola2017image}& 44.847&0.029$\pm$0.009\\
CycleGAN \cite{zhu2017unpaired}& 47.078&0.038$\pm$0.012\\
Her2 \cite{bai2022label}& 46.275&0.029$\pm$0.009\\
 RegGAN \cite{kong2021breaking}& 26.753&0.018$\pm$0.006\\
UNSB \cite{kim2023unpaired}& 25.557&0.012$\pm$0.005\\
 StegoGAN \cite{wu2024stegogan}& 73.590&0.053$\pm$0.009\\
 ASP \cite{adaptive_Supervised_patch}& 23.018&\pmb{0.010$\pm$0.006}\\
 ST \cite{stain_transformation}& 95.599&0.078$\pm$0.010\\
DGR& \pmb{20.264}&0.011$\pm$0.005\\
\hline
\end{tabular}
\end{table*}

\begin{table*}
\centering
\caption{Performance of different methods on staining the H\&E images to PAS-AB images. The mean and 95\% CI are reported.}
\label{tab:he2pas-ab}
\begin{tabular}{lllll}
\hline
 &Cohort&PSNR$\uparrow$&SSIM$\uparrow$&LPIPS$\downarrow$ \\
\hline
BCI \cite{Liu_2022_CVPR}&internal&17.670 (17.467-17.885)&0.488 (0.478-0.497)&0.203 (0.199-0.207) \\
Pix2pix \cite{isola2017image}&internal&17.459  (17.279-17.656)&0.452 (0.443-0.462)&0.206 (0.202-0.210) \\
CycleGAN \cite{zhu2017unpaired}&internal&13.336 (13.100-13.574)&0.378 (0.367-0.390)&0.268 (0.262-0.273) \\
Her2 \cite{bai2022label}&internal& 18.320 (18.155-18.499)& 0.534  (0.524-0.544)&0.180 (0.176-0.183) \\
RegGAN \cite{kong2021breaking}&internal& 18.455 (18.248-18.676)& 0.542 (0.532-0.552)&0.174 (0.170-0.177) \\
 UNSB \cite{kim2023unpaired}
& internal& 12.645 (12.433-12.863)& 0.350 (0.338-0.361)& 0.282 (0.278-0.287) \\
 StegoGAN \cite{wu2024stegogan}
& internal& 12.789 (12.581-13.009)&  0.364 (0.352-0.376)& 0.267 (0.262-0.271) \\
 ASP \cite{adaptive_Supervised_patch}
& internal& 13.238 (13.025-13.453)& 0.374 (0.363-0.386)&0.269 (0.265-0.274) \\
 ST \cite{stain_transformation}& internal&  16.307 (16.131-16.482)& 0.313 (0.301-0.325)&0.245 (0.242-0.249) \\
DGR&internal&\pmb{18.677 (18.481-18.872)}&\pmb{0.547  (0.537-0.557)}&\pmb{0.168 (0.165-0.172)} \\
\hline
 BCI \cite{Liu_2022_CVPR}& external& 13.406 (13.304-13.508)&  0.322 (0.317-0.326)&0.328  (0.323-0.332) \\
 Pix2pix \cite{isola2017image}& external& 15.011 (14.927-15.093)& 0.329 (0.325-0.333)& 0.301 (0.297-0.305) \\
 CycleGAN \cite{zhu2017unpaired}& external& 14.525 (14.386-14.658)& 0.465  (0.459-0.471)&0.245  (0.242-0.248) \\
 Her2 \cite{bai2022label}& external& 15.204 (15.106-15.306)& 0.461  (0.456-0.465)&0.262 (0.258-0.266) \\
 RegGAN \cite{kong2021breaking}& external& 15.409 (15.319-15.497)& 0.446 (0.441-0.451)&0.254  (0.250-0.258) \\
 UNSB \cite{kim2023unpaired}
& external& 14.601 (14.477-14.734)&  0.452 (0.446-0.459)& 0.247 (0.244-0.249)\\
 StegoGAN \cite{wu2024stegogan}
& external&  11.609 (11.471-11.745)&  0.135 (0.130-0.139)&  0.365 (0.360-0.370)\\
 ASP \cite{adaptive_Supervised_patch}
& external& 15.194 (15.073-15.318)& 0.452 (0.446-0.458)& 0.233 (0.230-0.235)\\
 ST \cite{stain_transformation}& external& 11.610 (11.473-11.737)& 0.135 (0.130-0.139)& 0.365 (0.360-0.369)\\
 DGR& external& \pmb{16.961 (16.908-17.013)}&  \pmb{0.502 (0.497-0.507)}&\pmb{0.220 (0.217-0.223)} \\
 \hline
\end{tabular}
\end{table*}
  
\begin{table*}
\centering
\caption{Performance of different methods on staining the H\&E images to PAS-AB images. 
Performance is measured using the Fréchet Inception Distance (FID) and the Kernel Inception Distance (KID, reported as mean $\pm$ standard deviation).
}
\label{tab:he2pas-ab}
\begin{tabular}{llll}
\hline
 &Cohort& FID$\downarrow$&KID$\downarrow$\\
\hline
BCI \cite{Liu_2022_CVPR}&internal& 37.687&0.016$\pm$0.004\\
Pix2pix \cite{isola2017image}&internal& 32.784&0.014$\pm$0.005\\
CycleGAN \cite{zhu2017unpaired}&internal& 23.743&0.003$\pm$0.003\\
Her2 \cite{bai2022label}&internal& 31.566&0.009$\pm$0.004\\
RegGAN \cite{kong2021breaking}&internal& 28.172&0.008$\pm$0.005\\
 UNSB \cite{kim2023unpaired}
& internal& 36.756&0.015$\pm$0.004\\
 StegoGAN \cite{wu2024stegogan}
& internal& 38.477&0.015$\pm$0.004\\
 ASP \cite{adaptive_Supervised_patch}
& internal& 35.557&0.014$\pm$0.004\\
 ST \cite{stain_transformation}& internal& 55.961&0.032$\pm$0.007\\
DGR&internal& \pmb{27.771}&\pmb{0.006$\pm$0.003}\\
\hline
 BCI \cite{Liu_2022_CVPR}& external& 55.191&0.035$\pm$0.006\\
 Pix2pix \cite{isola2017image}& external& 62.116&0.037$\pm$0.008\\
 CycleGAN \cite{zhu2017unpaired}& external& \pmb{29.992}&\pmb{0.014$\pm$0.003}\\
 Her2 \cite{bai2022label}& external& 54.280&0.031$\pm$0.007\\
 RegGAN \cite{kong2021breaking}& external& 55.168&0.031$\pm$0.007\\
 UNSB \cite{kim2023unpaired}
& external& 40.611&0.024$\pm$0.004\\
 StegoGAN \cite{wu2024stegogan}
& external& 42.554&0.026$\pm$0.005\\
 ASP \cite{adaptive_Supervised_patch}
& external& 40.898&0.024$\pm$0.005\\
 ST \cite{stain_transformation}& external& 100.688&0.069$\pm$0.008\\
 DGR& external& 49.231&0.027$\pm$0.006\\
 \hline
\end{tabular}
\end{table*}

  \begin{table*}[h]
\centering
\caption{Performance of different methods on translating H\&E images to mIHC images with HEMIT dataset. The mean and 95\% CI are reported.}
\label{tab:hemit}
\begin{tabular}{llll}
\hline
&PSNR$\uparrow$&SSIM$\uparrow$&LPIPS$\downarrow$ \\
\hline
BCI \cite{Liu_2022_CVPR}&25.142 (24.761-25.562)&0.752 (0.746-0.758)&0.414 (0.405-0.422) \\
Pix2pix  \cite{isola2017image}&25.011 (24.493-25.571)&0.736 (0.730-0.743)&0.395 (0.388-0.402) \\
CycleGAN \cite{zhu2017unpaired}&21.741 (21.504-22.014)&0.556 (0.549-0.562)&0.533 (0.525-0.540) \\
Her2 \cite{bai2022label}& 24.246 (23.842-24.637)& 0.750  (0.743-0.757)&0.393 (0.383-0.402) \\
RegGAN \cite{kong2021breaking}& 25.982 (25.672-26.281)& 0.810 (0.803-0.817)&0.258 (0.252-0.265) \\
 UNSB \cite{kim2023unpaired}
& 22.095 (21.901-22.291)& 0.579 (0.572-0.587)&0.500 (0.493-0.507) \\
 StegoGAN \cite{wu2024stegogan}
& 20.833 (20.565-21.125)&  0.575 (0.568-0.582)&0.549 (0.541-0.556) \\
 ASP \cite{adaptive_Supervised_patch}
&  25.114 (24.841-25.390)& 0.750 (0.742-0.757)&0.312 (0.306-0.319) \\
 ST \cite{stain_transformation}& 25.796 (25.411-26.182)&  0.783 (0.778-0.789)&0.317 (0.311-0.323) \\
DGR&\pmb{27.306 (26.939-27.694)}&\pmb{0.830 (0.826-0.834)}&\pmb{0.248 (0.242-0.253)} \\
\hline
\end{tabular}
\end{table*}

  \begin{table*}[h]
\centering
\caption{Performance of different methods on translating H\&E images to mIHC images with HEMIT dataset. 
Performance is measured using the Fréchet Inception Distance (FID) and the Kernel Inception Distance (KID, reported as mean $\pm$ standard deviation).
}
\label{tab:hemit_fid}
\begin{tabular}{lll}
\hline
&FID$\downarrow$&KID$\downarrow$\\
\hline
BCI \cite{Liu_2022_CVPR}&49.565&0.031$\pm$0.006\\
Pix2pix  \cite{isola2017image}&\pmb{33.184}&0.015$\pm$0.005\\
CycleGAN \cite{zhu2017unpaired}&113.133&0.121$\pm$0.021\\
Her2 \cite{bai2022label}& 81.818& 0.065$\pm$0.008\\
RegGAN \cite{kong2021breaking}& 36.194& \pmb{0.013$\pm$0.006}\\
 UNSB \cite{kim2023unpaired}
& 55.360& 0.034$\pm$0.010\\
 StegoGAN \cite{wu2024stegogan}
& 58.256&  0.032$\pm$0.010\\
 ASP \cite{adaptive_Supervised_patch}
&  67.676& 0.048$\pm$0.010\\
 ST \cite{stain_transformation}& 81.837&  0.057$\pm$0.008\\
DGR&56.083&0.034$\pm$0.006\\
\hline
\end{tabular}
\end{table*}

\begin{table*}[h]
\centering
\caption{Performance of different methods on stain normalization with Aperio2Hamamatsu dataset. The mean and CI are reported.}
\label{tab:aperio}
\begin{tabular}{llll}
\hline
&PSNR$\uparrow$&SSIM$\uparrow$&LPIPS$\downarrow$\\
\hline
BCI \cite{Liu_2022_CVPR}&16.492 (16.447-16.537)&0.260 (0.257-0.264& 0.193 (0.191-0.194)\\
Pix2pix \cite{isola2017image}&16.961 (16.909-17.014)&0.278 (0.274-0.282)&0.172 (0.171-0.173)\\
CycleGAN \cite{zhu2017unpaired}& 22.166 (22.116-22.218)&0.674 (0.671-0.676)& 0.119 (0.118-0.120)\\
Her2 \cite{bai2022label}& 17.695 (17.655-17.738)& 0.320 (0.318-0.323)&0.187 (0.186-0.187)\\
RegGAN \cite{kong2021breaking}&23.429 (23.367-23.491)&  0.719 (0.717-0.722)&\pmb{0.102 (0.101-0.103)}\\
 UNSB \cite{kim2023unpaired}
& 22.324 (22.263-22.384)& 0.688 (0.686-0.691)&0.130 (0.129-0.132)\\
 StegoGAN \cite{wu2024stegogan}
& 22.001 (21.928-22.078)& 0.692 (0.689-0.695)&0.124 (0.123-0.125)\\
 ASP \cite{adaptive_Supervised_patch}
& 22.396 (22.342-22.450)& 0.673 (0.671-0.675)&0.124 (0.123-0.125)\\
 ST \cite{stain_transformation}& 17.533 (17.491-17.576)& 0.270 (0.267-0.273)& 0.201 (0.200-0.202)\\
DGR&\pmb{23.823 (23.756-23.888)}&\pmb{0.734 (0.732-0.736)}&0.107 (0.106-0.108)\\
\hline
\end{tabular}
\end{table*}
\begin{table*}[h]
\centering
\caption{Performance of different methods on stain normalization with Aperio2Hamamatsu dataset. 
Performance is measured using the Fréchet Inception Distance (FID) and the Kernel Inception Distance (KID, reported as mean $\pm$ standard deviation).
}
\label{tab:aperio_fid}
\begin{tabular}{lll}
\hline
&FID$\downarrow$&KID$\downarrow$\\
\hline
BCI \cite{Liu_2022_CVPR}&39.797&0.037$\pm$0.006\\
Pix2pix \cite{isola2017image}&20.404&0.019$\pm$0.005\\
CycleGAN \cite{zhu2017unpaired}& 10.832&0.008$\pm$0.004\\
Her2 \cite{bai2022label}& 23.905& 0.019$\pm$0.004\\
RegGAN \cite{kong2021breaking}&12.235&  0.009$\pm$0.005\\
 UNSB \cite{kim2023unpaired}
& 16.894& 0.013$\pm$0.005\\
 StegoGAN \cite{wu2024stegogan}
& 19.925& 0.018$\pm$0.004\\
 ASP \cite{adaptive_Supervised_patch}
& 17.877& 0.015$\pm$0.004\\
 ST \cite{stain_transformation}& 91.607& 0.095$\pm$0.010\\
DGR&\pmb{10.253}&\pmb{0.007$\pm$0.004}\\
\hline
\end{tabular}
\end{table*}

\begin{figure*}[h]
    \centering
    \includegraphics[width=1\linewidth]{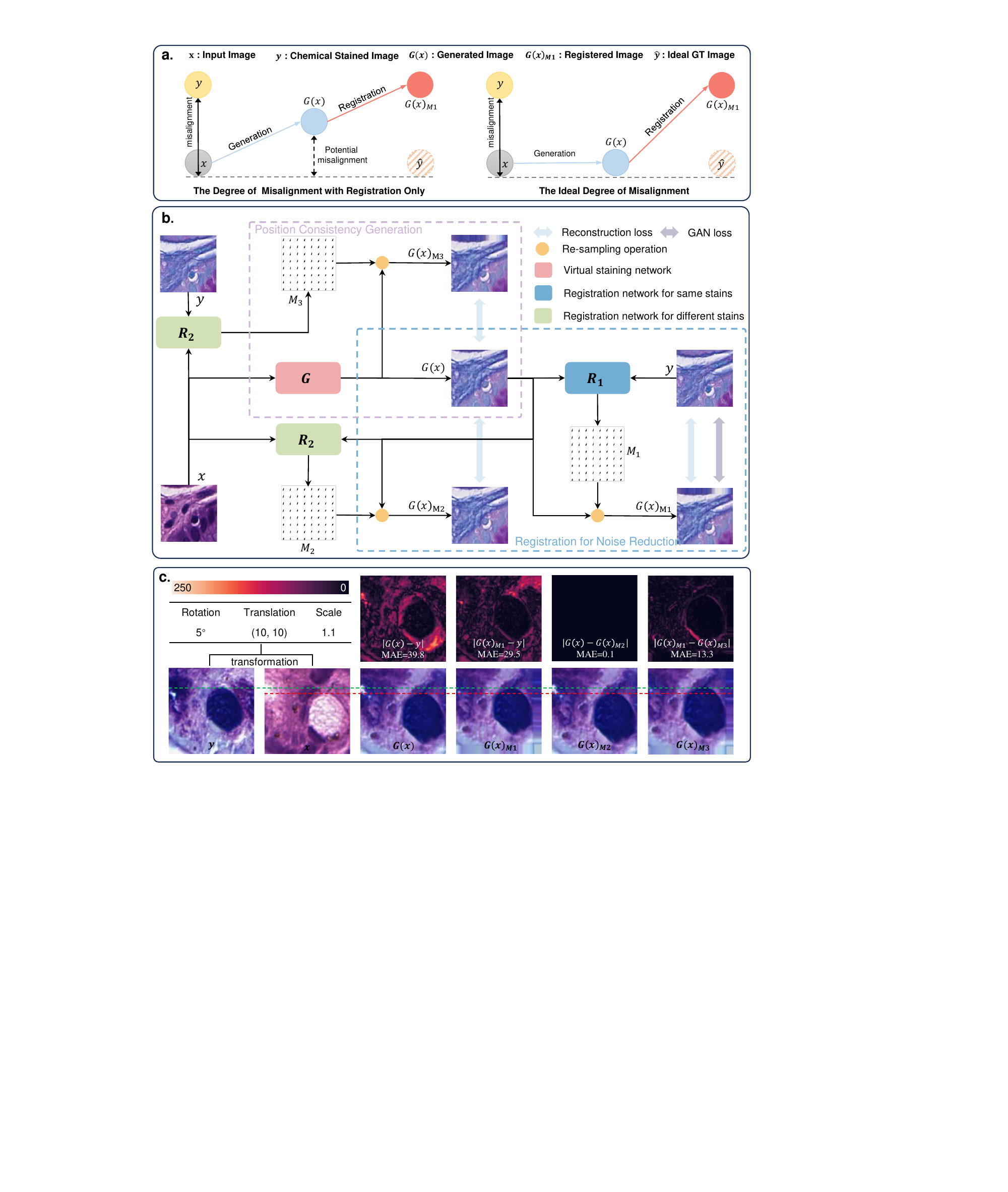}
    \caption{\textbf{Architecture and Visualization of DGR}. 
    \textbf{a}. Illustration of the misalignment between the input image, generated image, and registered image.     \textbf{b}. The DGR framework consists of two key components: the registration module and the position consistency module. Here, $M_i$ represents the deformation field produced by the registration network. 
    \textbf{c}. Demonstration of DGR's resilience to misalignment across varying levels. The table specifies the misalignment levels, and the corresponding virtually stained images and registered images are displayed. Notably, the models used to generate these images were trained on data with level 3 misalignment. 
    }
    \label{fig:DTR-arch}
\end{figure*}
\begin{figure*}[h]
    \centering
    \includegraphics[width=1\linewidth]{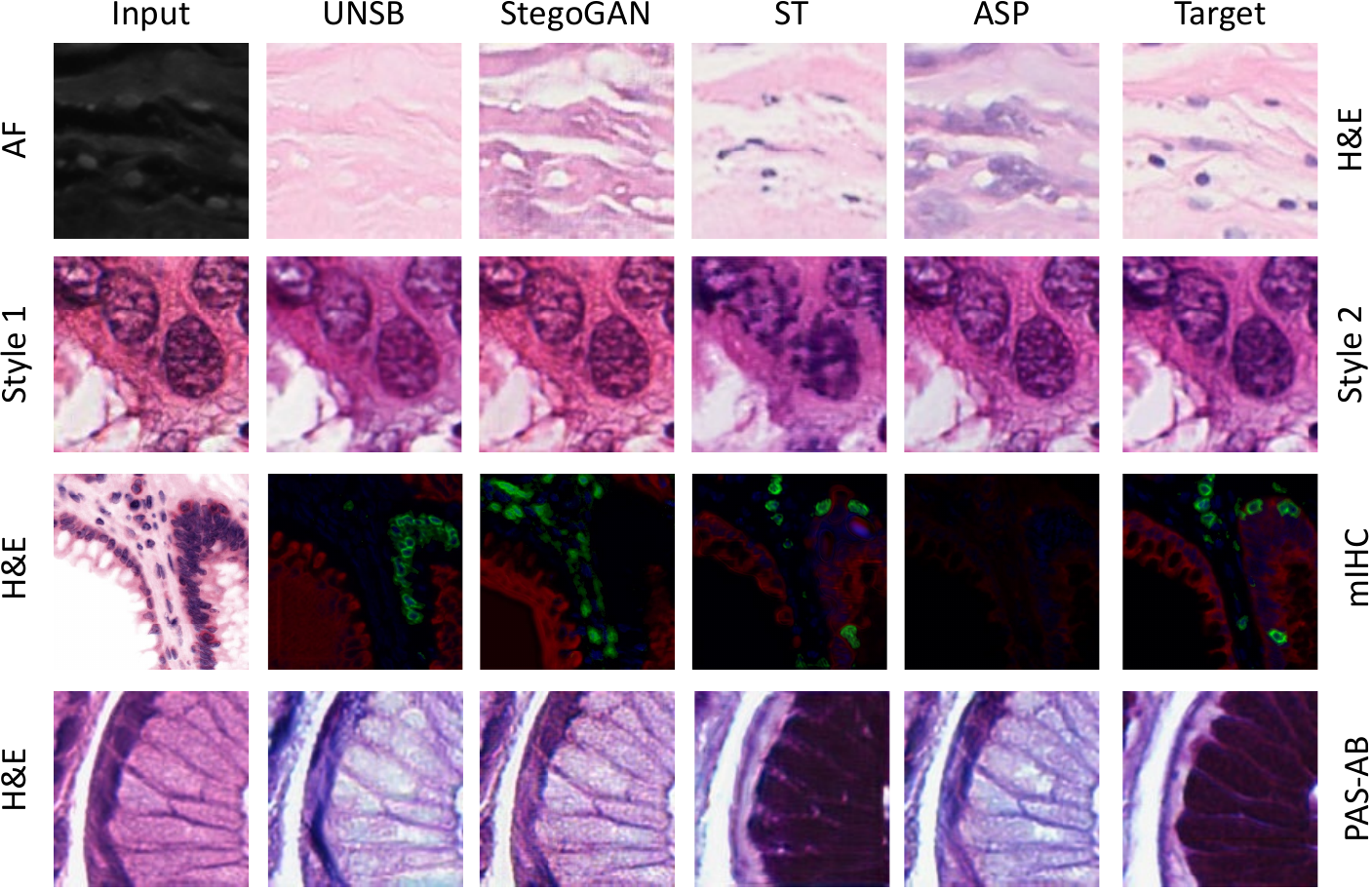}
\caption{Visualization of virtually stained images by four different methods: ASP, UNSB, ST, and StegoGAN.
From top to bottom, each row corresponds to the AF2HE, A2H, HEMIT, and HE2PAS-AB datasets, respectively. 
}
    \label{fig:extra_results}
\end{figure*}

\begin{figure*}[h]
    \centering
    \includegraphics[width=1\linewidth]{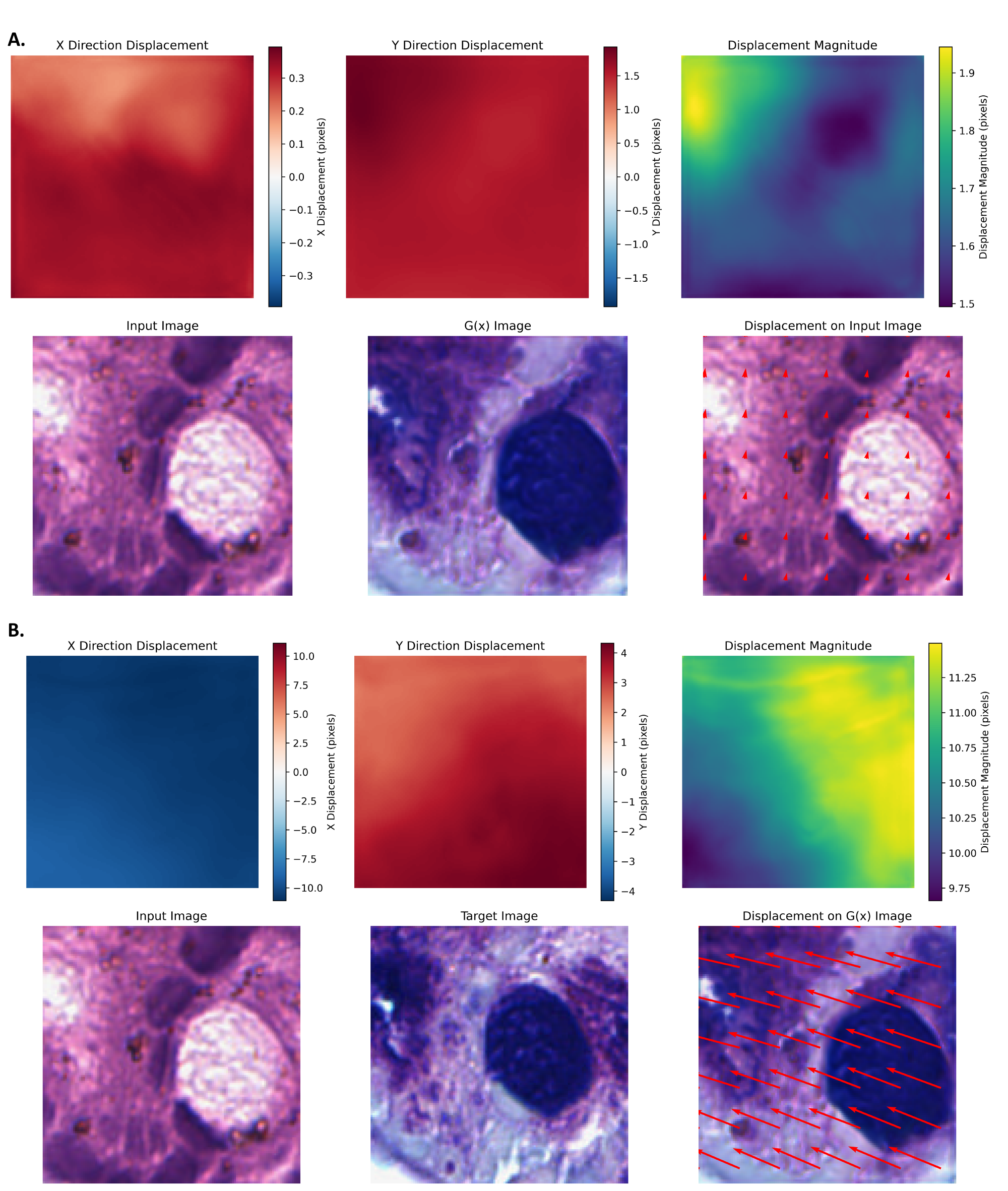}
\caption{Visualization of the deformation fields learned by the DGR framework. 
\textbf{a.} Deformation field between the input H\&E image and the generated image $G(x)$.
\textbf{b.} Deformation field between $G(x)$ and the target ground truth image, generated by model $R_1$ (see Extended Data Figure \ref{fig:DTR-arch}).
The strong alignment in subfigure a. shows $G(x)$ is a faithful transformation of the input, which is enhanced by the registration model R2. The effective alignment in subfigure b. demonstrates the registration mechanism of $R_1$ is a core and effective component of the DGR framework.
}
    \label{fig:deformation_field}
\end{figure*}

\begin{figure*}
    \centering
    \includegraphics[width=1\linewidth]{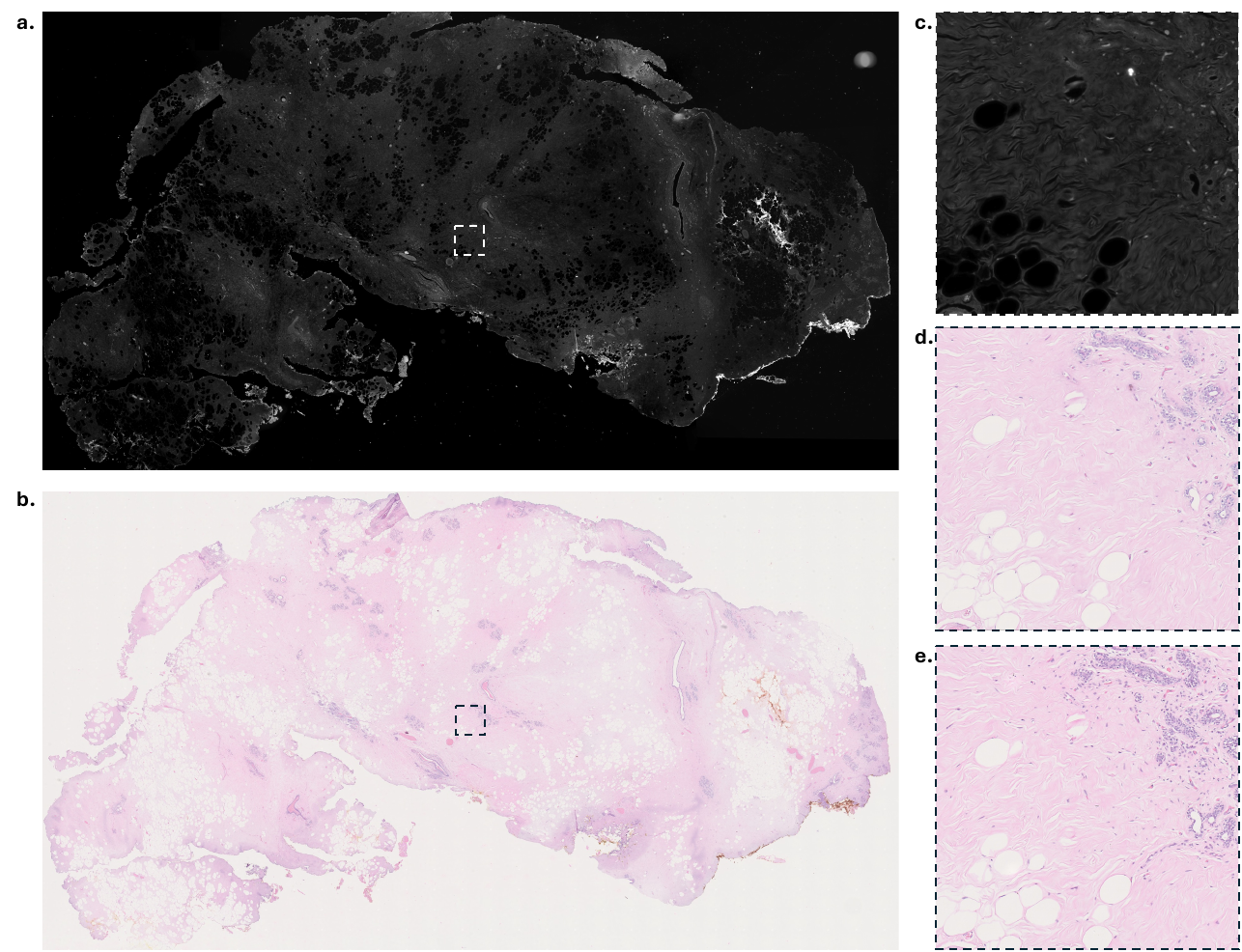}
    \caption{\textbf{The Whole Slide Image (WSI)Visualization on virtual staining autofluorescence (AF) images to H\&E images.} \textbf{a.} The H\&E WSI (input). \textbf{b.} The H\&E WSI (GT). \textbf{c.} The zoomed-in autofluorescence image. \textbf{d.} The virtual stained large H\&E image by our model, is stitched by 256 virtual stained patches. \textbf{e.} The zoomed-in H\&E GT.}
    \label{fig:af2hewsi}
    
\end{figure*}

\begin{figure*}
    \centering
    \includegraphics[width=1\linewidth]{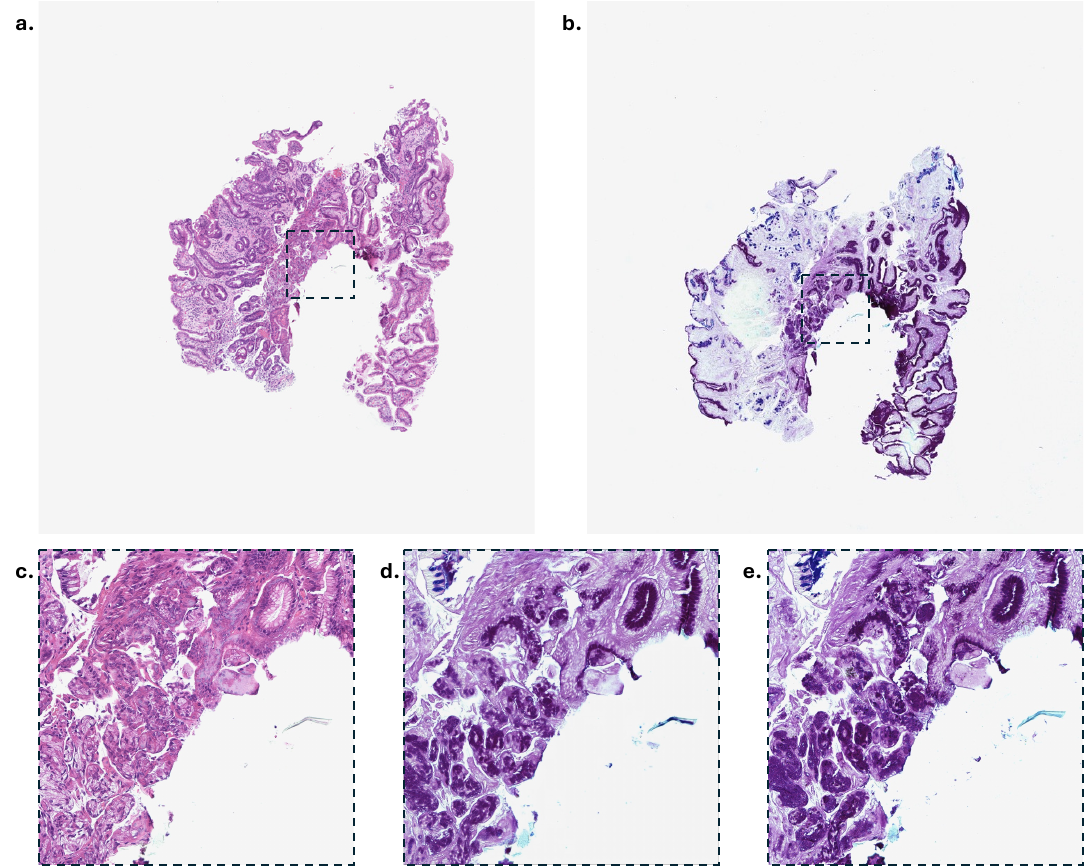}
    \caption{\textbf{The Whole Slide Image (WSI) Visualization on virtual staining H\&E images to PAS-AB images.} \textbf{a.} The autofluorescence WSI (input). \textbf{b.} The PAS-AB WSI (GT). \textbf{c.} The zoomed-in H\&E image. \textbf{d.} The virtual stained large PAS-AB image by our model, is stitched by 256 virtual stained patches. \textbf{e.} The zoomed-in PAS-AB GT.
    }
    \label{fig:he2paswsi}
    
\end{figure*}
\textbf{Cellular Structure Analysis} We employ Cellpose \cite{stringer2021cellpose,pachitariu2025cellpose} to segment cell nuclei, using the chemically stained HE2PAB-AS dataset as the ground truth. The same methodology is applied to segment cells in virtually stained images generated by different models. We then compute Dice coefficient (DICE) to quantitatively evaluate segmentation accuracy. The model whose results most closely align with the ground truth HE2PAS-AB data is considered superior in preserving cellular structural similarity. 
\begin{table*}[h]
\centering
\caption{Quantitative comparison of cellular structure preservation using Cellpose-based segmentation. Higher DICE values indicate better structural consistency with the ground truth HE2PAS-AB dataset }
\label{tab:cellpose}
\begin{tabular}{ll}
\hline
&DICE\\
\hline
BCI \cite{Liu_2022_CVPR}&0.396 \\
Pix2pix \cite{isola2017image}&0.363\\
CycleGAN \cite{zhu2017unpaired}&0.265 \\
Her2 \cite{bai2022label}&0.397 \\
RegGAN \cite{kong2021breaking} & 0.421\\
 UNSB \cite{kim2023unpaired}
&0.291\\
 StegoGAN \cite{wu2024stegogan}
&0.303\\
 ASP \cite{adaptive_Supervised_patch}
&0.290\\
 ST \cite{stain_transformation}&0.270 \\
 DGR&\pmb{0.422}\\
\hline
\end{tabular}
\end{table*}

\begin{figure*}
        \centering
    \includegraphics[width=1\linewidth]{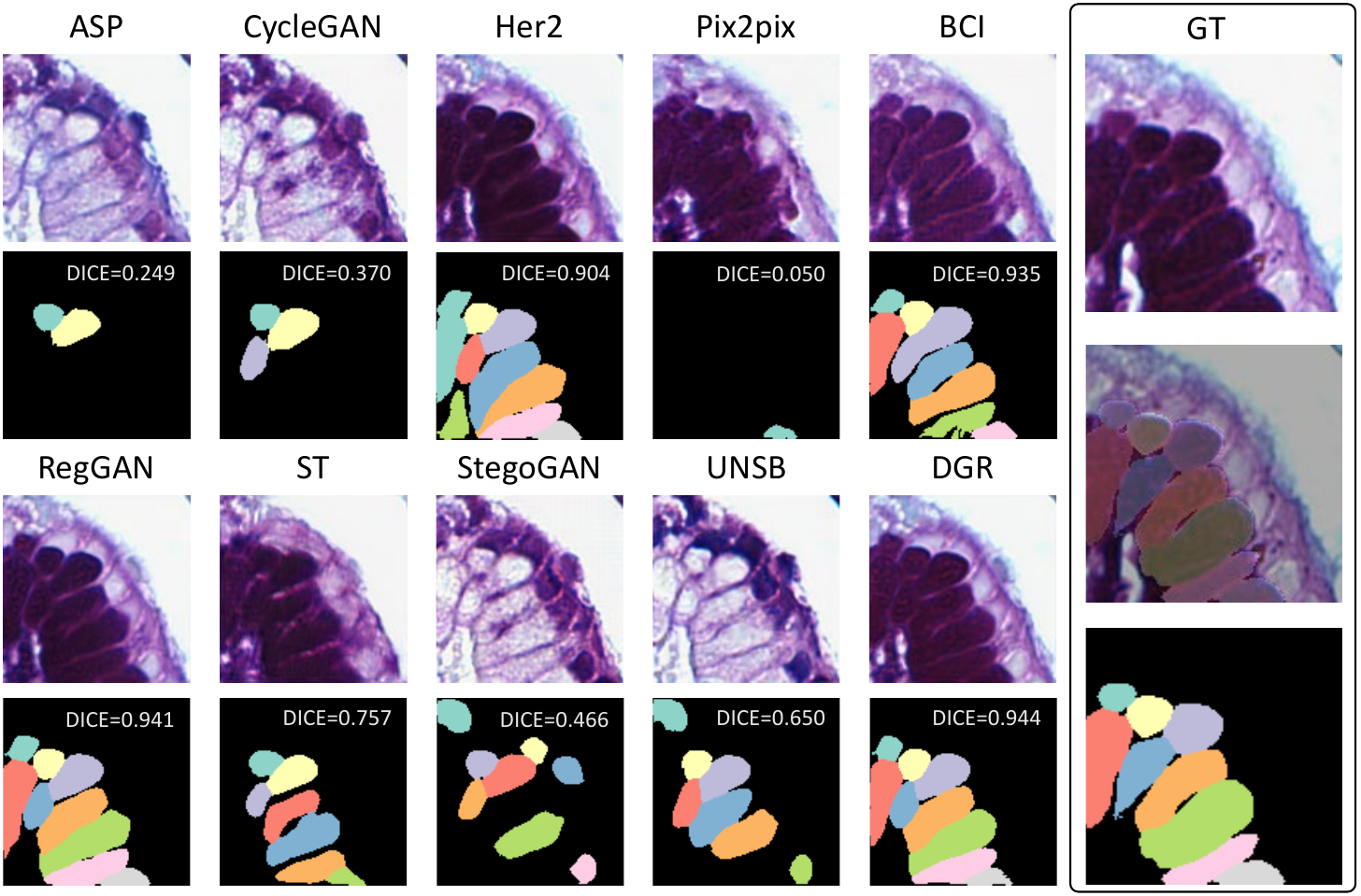}
\caption{Cellular structure segmentation of generated and ground truth (GT) images. The segmentation was performed using Cellpose. The Dice coefficient (DICE) was calculated by comparing the generated image mask against the GT mask as the reference standard.}
    \label{fig:cellpose}
\end{figure*}

\begin{sidewaystable*}[h]
    \centering
    \caption{The misalignment resilience of DGR on different level of misalignment.
Note that the model is trained on misaligned data and tested on well-aligned data. 
Baseline is the Bai et al. \cite{bai2022label} model.
RNR indicates the registration for Noise Reduction module. 
PcG denotes the Position-consistency Generation module.
    }
    \label{ablation_study}
    \begin{tabular}{lllllccccc}
    \hline
     &     Baseline&RNR&PcG&& Level 1& Level 2& Level 3& Level 4& Level 5\\ 
    \hline
    \multicolumn{4}{l}{\multirow{3}{*}{Misalignment Setting}}&Rotation & ±1° & ±2° & ±3° & ±4° & ±5° \\
     &    &&&Translation & ±2\%& ±4\%& ±6\%& ±8\%& ±10\%\\
     &    &&&Rescaling & ±2\%& ±4\%& ±6\%& ±8\%& ±10\%\\ \hline
    \multirow{3}{*}{Exp. 1}&\multirow{3}{*}{\checkmark}&&&SSIM&  0.412±0.172&  0.205±0.136&  0.301±0.191&  0.223±0.184&  0.226±0.202\\
     &     &&&PSNR&   17.702±2.799&  16.109±2.303&  15.694±3.407&  15.818±2.373&  14.324±3.239\\
     &     &&&LPIPS&  0.205±0.062&   0.277±0.059&  0.267±0.065&  0.390±0.068&  0.330±0.065\\ \hline
    \multirow{3}{*}{Exp. 2}&     \multirow{3}{*}{\checkmark}&\multirow{3}{*}{\checkmark}&&SSIM
&  0.522±0.175&  0.507±0.177&  0.511±0.170&  0.493±0.171&   0.430±0.175\\
     &     &&&PSNR
&  18.228±3.584&  17.988±3.490&   18.031±3.443&  17.780±3.287&  17.095±3.647\\
     &     &&&LPIPS&   0.176±0.064&   0.181±0.064&   0.179±0.064&   0.183±0.064&  0.206±0.065\\ \hline
    \multirow{3}{*}{Exp. 3}&     \multirow{3}{*}{\checkmark}&\multirow{3}{*}{\checkmark}&\multirow{3}{*}{\checkmark}&SSIM
&  \pmb{0.528±0.171}&  \pmb{0.532±0.174}&  \pmb{0.530±0.173}&  \pmb{0.512±0.175}&  \pmb{0.466±0.174}\\
     &     &&&PSNR
&  \pmb{18.414±3.052}&  \pmb{18.393±3.447}&  \pmb{18.460±3.716}&\pmb{18.164±3.220}&\pmb{17.730±3.528}\\
     &     &&&LPIPS&\pmb{0.172±0.064}&\pmb{0.173±0.065}&\pmb{0.174±0.065}&\pmb{0.180±0.064}&\pmb{0.184±0.064}\\ 
     \hline
    \end{tabular}
\end{sidewaystable*}

\end{appendices}
\end{document}